\documentclass[runningheads]{llncs}
\usepackage{graphicx}
\usepackage{comment} 
\usepackage{amsmath,amssymb} %

\usepackage{xspace}

\usepackage{url}            %
\usepackage{booktabs}       %
\usepackage{amsfonts}       %
\usepackage{nicefrac}       %
\usepackage{microtype}      %

\usepackage{cuted}

\usepackage{enumitem}
\usepackage{bm}
\usepackage{rotating}
\usepackage{multirow}
\usepackage[percent]{overpic}

\usepackage[table,xcdraw]{xcolor}

\usepackage[width=122mm,left=12mm,paperwidth=146mm,height=193mm,top=12mm,paperheight=217mm]{geometry}

\usepackage{microtype}
\renewcommand{\vec}[1]{{\bm{#1}}}

\newcommand{\iminput}{\mathcal{A}}
\newcommand{\imcat}{\mathcal{C}}
\newcommand{\transform}{\mathcal{T}}

\newcommand{\winput}{w^{\iminput}}
\newcommand{\hinput}{h^{\iminput}}
\newcommand{\wcat}{w^{\imcat}}
\newcommand{\hcat}{h^{\imcat}}
\newcommand{\wtransform}{w^{\transform}}
\newcommand{\htransform}{h^{\transform}}

\newcommand{\ipos}{\mbox{\scriptsize pos}}
\newcommand{\ineg}{\mbox{\scriptsize neg}}
\newcommand{\irec}{\mbox{\scriptsize rec}}
\newcommand{\iloc}{\mbox{\scriptsize loc}}

\newcommand{\methodname}{\text{OS2D}\xspace}

\usepackage[pagebackref=true,breaklinks=true,colorlinks,bookmarks=false]{hyperref}

\definecolor{mydarkblue}{rgb}{0,0.08,0.45}

\hypersetup{ %
    pdftitle={OS2D: One-Stage One-Shot Object Detection by Matching Anchor Features},
    pdfauthor={Anton Osokin, Denis Sumin, Vasily Lomakin},
    pdfsubject={OS2D: One-Stage One-Shot Object Detection by Matching Anchor Features},
    pdfkeywords={structured prediction, consistency, surrogate losses, machine learning},
    pdfborder=0 0 0,
    pdfpagemode=UseNone,
    colorlinks=true,
    linkcolor=mydarkblue,
    citecolor=mydarkblue,
    filecolor=mydarkblue,
    urlcolor=mydarkblue,
    pdfview=FitH
}

\begin{document}
\pagestyle{headings}
\mainmatter
\def\ECCVSubNumber{2424}  %

\title{OS2D: One-Stage One-Shot Object Detection by Matching Anchor Features} %

\titlerunning{OS2D: One-Stage One-Shot Object Detection by Matching Anchor Features}

\author{Anton Osokin\inst{1,2} \and
Denis Sumin\inst{3} \and
Vasily Lomakin\inst{3}}

\authorrunning{A. Osokin, D. Sumin, V. Lomakin}

\institute{National Research University Higher School of Economics\footnote{This work was done when Anton Osokin was with the Samsung-HSE lab.}, Moscow, Russia
    \and Yandex, Moscow, Russia
    \and mirum.io, Moscow, Russia
}
\maketitle

\begin{abstract}
In this paper, we consider the task of one-shot object detection, which consists in detecting objects defined by a single demonstration.
Differently from the standard object detection, the classes of objects used for training and testing do not overlap. We build the one-stage system that performs localization and recognition jointly.
We use dense correlation matching of learned local features to find correspondences, a feed-forward geometric transformation model to align features and bilinear resampling of the correlation tensor to compute the detection score of the aligned features.
All the components are differentiable, which allows end-to-end training.
Experimental evaluation on several challenging domains (retail products, 3D objects, buildings and logos) shows that our method can detect unseen classes (e.g., toothpaste when trained on groceries) and outperforms several baselines by a significant margin. Our code is available online: \url{https://github.com/aosokin/os2d}
\keywords{one-shot detection, object detection, few-shot learning}
\end{abstract}

\section{Introduction}
The problem of detecting and classifying objects in images is often a necessary component of automatic image analysis.
Currently, the state-of-the-art approach to this task consists in training convolutional neural networks (CNNs) on large annotated datasets.
Collecting and annotating such datasets is often a major cost of deploying such systems and is the bottleneck when the list of classes of interest is large or changes over time.
For example, in the domain of retail products on supermarket shelves, the assortment of available products and their appearance is gradually changing (e.g., 10\% of products can change each month), which makes it hard to collect and maintain a dataset.
Relaxing the requirement of a large annotated training set will make the technology easier to apply.

\begin{figure}
\centering
    \begin{tabular}{@{}m{0.2\textwidth}@{}m{0.4\textwidth}@{}m{0.4\textwidth}@{}}
                \begin{minipage}{0.2\textwidth}%
                \begin{overpic}[width=\textwidth]{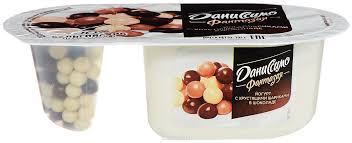}
                    \put (0,0){\colorbox{white}{\scriptsize 9}}
                \end{overpic}
                \begin{overpic}[width=\textwidth]{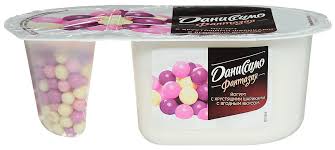}
                    \put (0,0){\colorbox{white}{\scriptsize 10}}
                \end{overpic}%
                \end{minipage}%
                 &
                 \includegraphics[width=0.37\textwidth, clip=true, trim=30mm 28mm 0mm 
                 15mm]{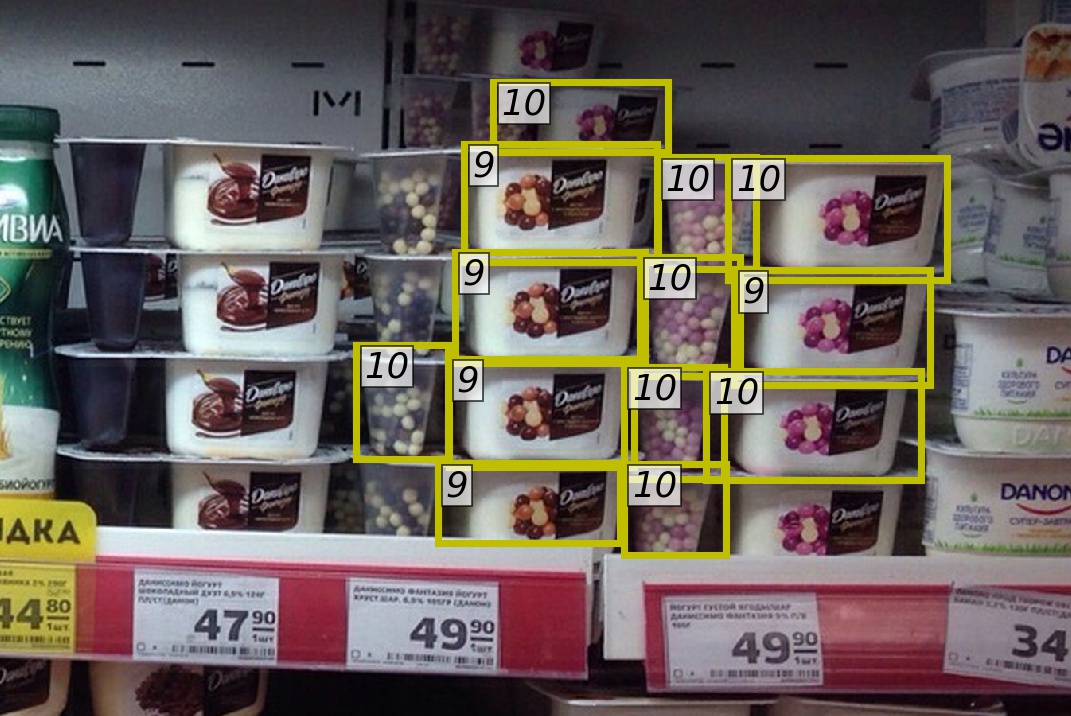}
                 &
                 \includegraphics[width=0.37\textwidth, clip=true, trim=30mm 28mm 0mm 15mm]{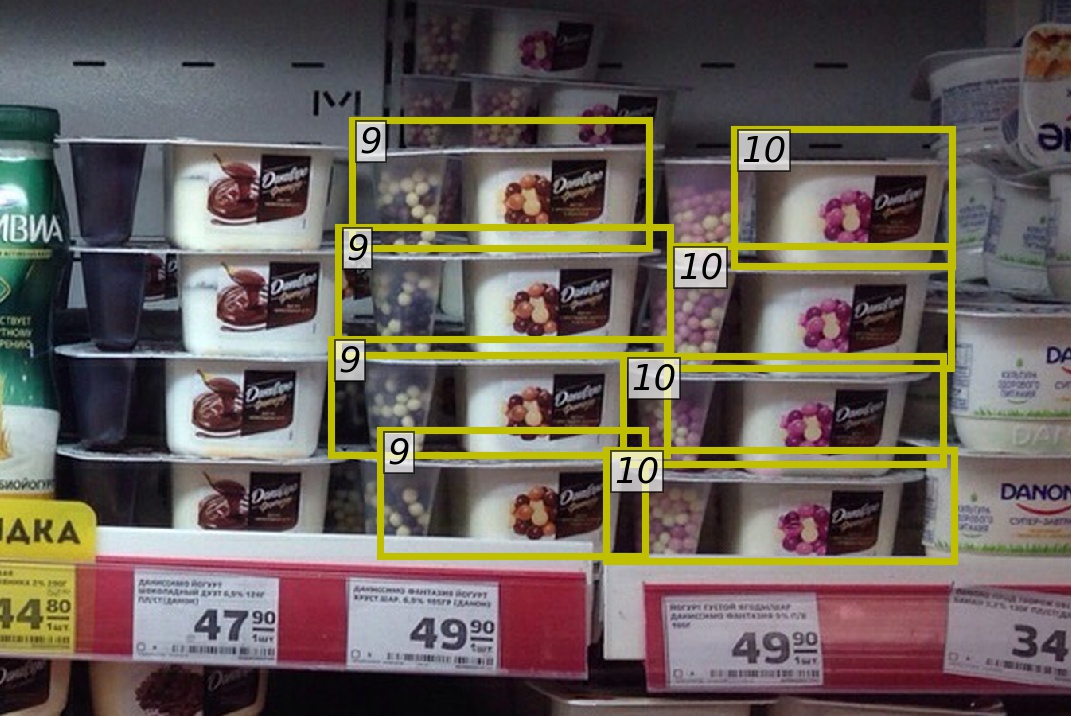}
                 \\
                 (a) classes &  (b) baseline & (c) ours: \methodname
                 \\[-2mm]
            \end{tabular}
    \caption{Qualitative comparison of our \methodname model with the baseline (state-of-the-art object detection and image retrieval systems) when detecting unseen object classes.
        In this example, the goal is to detect objects that consist of several parts (a), which were not available in the training set.
        The baseline system has to finalize the object bounding box before knowing what class it is detecting (b), so as the object detector fails to merge parts, the retrieval system cannot fix the boxes and recognize correctly~(c).
        Our model knows that the target objects consist of two parts, so it detects correctly
        \label{fig:teaser}}
\end{figure}

In this paper, we consider the task of detecting objects defined by a single demonstration (one-shot object detection).
In this formulation, a system at the detection (testing) stage has to ``understand'' what it needs to detect by a single exemplar of a target class.
After the demonstration, the system should detect objects of the demonstrated target class on new images.
Differently from the standard object detection systems, the classes of objects used for training and testing do not overlap.
The task of one-shot detection is a step towards decreasing the complexity of collecting and maintaining data for object detection systems.

The computer vision community developed a few very successful methods for regular object detection, and currently, there are well-maintained implementations in all major deep learning libraries.
One-shot detection is studied substantially less with one notable exception: human face detection combined with personality recognition.
A significant simplification of this setting is that a human face is a well-defined object, and one can build a detector that works well on unseen faces.
Defining a general object is harder~\cite{alexe2010object}, so such detector suffers from insufficient generalization.
Several recent works tackled one-shot detection for general object classes (ImageNet~\cite{ILSVRC15} and COCO~\cite{lin2014coco}).
Usually, methods have two distinct stages: a detector of all objects (a region proposal network) and an embedding block for recognition.
Works~\cite{karlinsky2018,wang2019,yan2019,wang2020} linked the two stages by joint finetuning of network weights but, at the test stage, their models relied on class-agnostic object proposals (no dependence on the target class image).
Differently, works~\cite{hsieh2019coae,fan2020} incorporated information about the target class into object proposals.

\textbf{Contributions}.
First, we build a one-stage one-shot detector, \methodname, by bringing together the dense grid of anchor locations of the Faster R-CNN~\cite{ren2015faster} and SSD~\cite{liu2016ssd} object detectors, the transformation model of Rocco et~al.~\cite{rocco2017,rocco2018weakalign} designed for semantic alignment and the differentiable bilinear interpolation~\cite{jaderberg2015}.
Our approach resembles the classical pipelines based on descriptor matching and subsequent geometric verification~\cite{lowe1999,lowe2004sift}, but utilizes dense instead of sparse matches, learned instead of hand-crafted local features and a feed-forward geometric transformation model instead of RANSAC.
The key feature of our model is that detection and recognition are performed jointly without the need to define a general object (instead, we need a good feature descriptor and transformation model).
Figure~\ref{fig:teaser} shows how detection and recognition can work jointly compared to the baseline consisting of separate detection and retrieval systems (see Section~\ref{sec:exp:baselines} for the baseline details).

Second, we design a training objective for our model that combines the ranked list loss~\cite{wang2019rll} for deep metric learning and the standard robust $L_1$ loss for bounding box regression~\cite{girshick2015fast}.
Our objective also includes remapping of the recognition targets based on the output of the forward pass, which allows us to use a relatively small number of anchors.

Finally, we apply our model to several domains (retail products based on the GroZi-3.2k dataset~\cite{george2014grozi}, everyday 3D objects, buildings and logos based on the INSTRE dataset~\cite{wang2015instre}).
For the retails products, we have created a new consistent annotation of all objects in GroZi-3.2k and collect extra test sets with new classes (e.g., toothpaste, when trained on groceries).
Our method outperformed several baselines by a significant margin in all settings.
Code of our method and the baselines together with all collected data is available online.

We proceed by reviewing the background works of Rocco et~al.~\cite{rocco2017,rocco2018pami,rocco2018weakalign} in Section~\ref{sec:matchingNetworks}.
We present our model and its training procedure in Sections~\ref{sec:model} and~\ref{sec:training}, respectively.
Next, in Section~\ref{sec:relatedworks}, we review the related works.
Finally, Section~\ref{sec:experiments} contains the experimental evaluation, Section~\ref{sec:conclusion}~-- the conclusion.

\section{Preliminaries: matching networks \label{sec:matchingNetworks}}
We build on the works of Rocco et~al.~\cite{rocco2017,rocco2018pami,rocco2018weakalign} that targeted the problem of semantic alignment where the goal was to align the source and target images.
Their methods operate on dense feature maps $\{\vec{f}_{kl}\}$, $\{\vec{g}_{kl}\}$ of the spatial size $\htransform \times \wtransform$ and feature dimensionality $d$.
The feature maps are extracted from both images with a ConvNet, e.g., VGG or ResNet.
To match the features, they compute a 4D tensor $c \in \mathbb{R}^{\htransform \!\!\times \wtransform \!\!\times \htransform \!\!\times \wtransform}\!\!$ containing correlations
$c[k,l,p,q] = \frac{ \langle \vec{f}_{kl}, \vec{g}_{pq} \rangle}{\|\vec{f}_{kl}\|_2 \|\vec{g}_{pq} \|_2}$ between all pairs of feature vectors from the two feature maps.

Next, they reshape the tensor~$c$ into a 3D tensor $\tilde{c} \in \mathbb{R}^{\htransform \!\!\times \wtransform \!\!\times (\htransform \!\wtransform)}$ and feed it into a ConvNet (with regular 2D convolutions) that outputs the parameters of the transformation aiming to map coordinates from the target image to the source image.
Importantly, the kernel of the first convolution of such ConvNet has $\htransform \!\wtransform$ input channels, and the overall network is designed to reduce the spatial size $\htransform \!\!\times \wtransform$ to $1 \times 1$, thus providing a single vector of transformation parameters.
This ConvNet will be the central element of our model, and hereinafter we will refer to it as TransformNet.
Importantly, TransformNet has 3 conv2d layers with ReLU and BatchNorm~\cite{ioffe2015batchnorm} in-between and has receptive field of $15 \times 15$, i.e., $\htransform = \wtransform = 15$  (corresponds to the image size $240 \times 240$ if using the features after the fourth block of ResNet).
The full architecture is given in Appendix~\ref{sec:transformnet_arch}.

TransformNet was first trained with full supervision on the level of points to be matched~\cite{rocco2017,rocco2018pami} (on real images, but synthetic transformations).
Later, Rocco et~al.~\cite[Section 3.2]{rocco2018weakalign} proposed a \emph{soft-inlier count} to finetune TransformNet without annotated point matches, i.e., with weak supervision.
The soft-inlier count is a differentiable function that evaluates how well the current transformation aligns the two images.
It operates by multiplying the tensor of correlations~$c$ by the soft inlier mask, which is obtained by warping the identity mask with a spatial transformer layer~\cite{jaderberg2015} consisting of grid generation and bilinear resampling.

\section{The \methodname model \label{sec:model}}
We now describe our \methodname model for one-shot detection.
The main idea behind \methodname is to apply TransformNet to a large feature map in a fully-convolutional way and to modify the soft-inlier count to work as the detection score.
We will explain how to modify all the components of the model for our setting.

\begin{figure}[t]
    \centering
        \includegraphics[width=\textwidth,clip=true,trim=0mm 0mm 113mm 0mm]{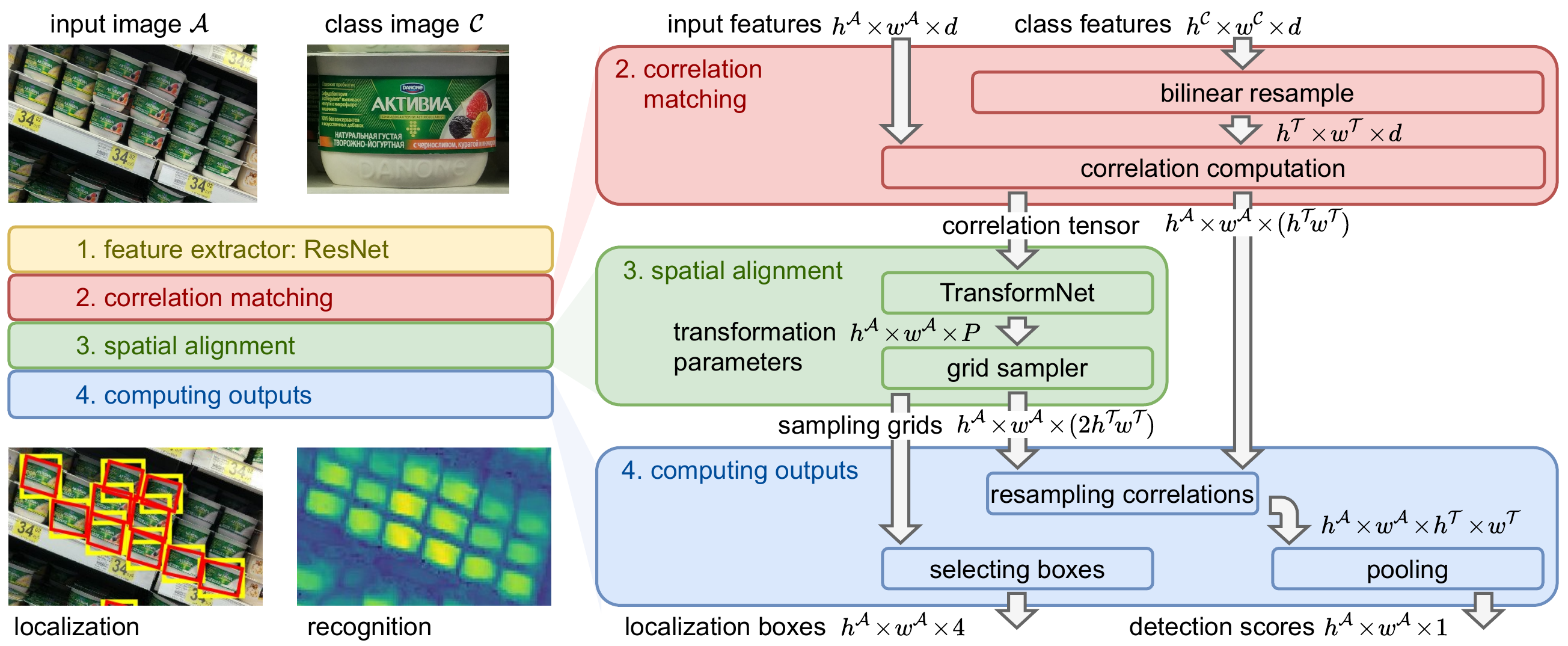}
    \caption{\textbf{(left)} Inputs, outputs and main components of the \methodname model.
        Bottom left~-- the detection boxes (yellow) that correspond to the peaks in the score map, red parallelograms illustrate the corresponding affine transformations produced by TransformNet; and the detection score map (the lighter~-- the higher the score).
        \textbf{(right)} Details of the \methodname model.
        Boxes represent network layers, arrows~-- data flow (for convenience, we show the sizes of  intermediate tensors).
        Best viewed in color
        \label{fig:model}}
\end{figure}

The \methodname model consists of the following steps (see Figure~\ref{fig:model}):
(1) extracting local features from both input  and class images;
(2) correlation matching of features;
(3) spatially aligning features according to successful matches;
(4) computing the localization bounding boxes  and recognition score.
We now describe these steps highlighting the technical differences to~\cite{rocco2017,rocco2018pami,rocco2018weakalign}.

\textbf{Feature extraction.}
On an input image~$\iminput$ and a class image~$\imcat$, a feature extractor (same as in  works~\cite{rocco2017,rocco2018pami,rocco2018weakalign,rocco2018consensus}) computes dense feature maps $\{\vec{f}^{\iminput}_{kl}\} \in \mathbb{R}^{\hinput \times \winput \times d}$ and $\{\vec{f}^{\imcat}_{pq}\} \in \mathbb{R}^{\hcat \times \wcat \times d}$, respectively.
The spatial sizes of the two feature maps differ and depend linearly on the sizes of the respective images.

The architecture of TransformNet requires at least one of the extracted feature maps to be of the fixed size $\htransform \!\!\times \wtransform$ 
(this size defines the number of the input channels of the first TransformNet convolution).
As the input image~$\iminput$ can be of a large resolution and thus need a large feature map, we chose to convert the class feature map~$\{\vec{f}^{\imcat}_{pq}\}$ to the fixed size to get~$\{\vec{f}^{\transform}_{pq}\}$.
Our experiments showed that resizing class images (as in \cite{rocco2017,rocco2018pami,rocco2018weakalign}) significantly distorts the aspect ratio of some classes, e.g., a bottle becomes a can, which degrades the quality of feature matching.
We found that it worked better to resize extracted class feature maps directly by applying the differentiable bilinear resampling~\cite{jaderberg2015}.

We also found that it is very important to extract features from the input and class images in the Siamese way, i.e., with the networks with identical parameters, which ruled out more efficient (than an image pyramid) ways of computing multi-scale features maps, e.g., FPN~\cite{lin2017fpn}.
We tried to untie the two branches and use FPN to extract features from the input image: such system worked well on the classes seen during training, but did not generalize to new classes.

\textbf{Feature matching and alignment.}
Following Rocco et~al.~\cite{rocco2017} we compute a 4D correlation tensor $c \in \mathbb{R}^{\hinput \!\times \winput \!\times \htransform \!\times \wtransform}\!\!$, $c[k,l,p,q] = \frac{ \langle \vec{f}^{\iminput}_{kl}, \vec{f}^{\transform}_{pq} \rangle}{\|\vec{f}^{\iminput}_{kl}\|_2 \|\vec{f}^{\transform}_{pq} \|_2}$, between the input and resized class feature maps.
After reshaping the tensor~$c$ to the 3D tensor $\tilde{c} \in \mathbb{R}^{\hinput \!\times \winput \!\times (\htransform \wtransform)}$, we apply TransformNet in a fully-convolutional way to obtain a $\hinput \!\times \winput \!\times P$ tensor with parameters of transformations at each location of the input feature map.
All transformations are defined w.r.t.\ local coordinates (different for all locations), which we then convert to the global w.r.t. the input feature map coordinates (we denote the transformations by $G_{kl}$).

The direction of the output transformations $G_{kl}$ is defined by training the model.
The TransformNet weights released by Rocco et~al.~\cite{rocco2018weakalign} align the input image to the class image (according to their definition of the source and target), but we need the inverse of this to map everything to the coordinates w.r.t.\ the input image.
Note that we cannot simply swap the two images because their feature maps are of different sizes.
To use the released weights, we need to invert the direction, which, in the case of affine transformations, requires a batch inversion of $3\times 3$ matrices.
An alternative is to retrain TransformNet from scratch and to interpret its output transformations in the direction we need.

Finally, we feed all the transformations $G_{kl}$ into a grid sampler to produce a grid of points aligning the class image  at each location of the input image (we use the grids of the same size $\htransform \!\times \wtransform$ as the TransformNet input).
The output of the grid sampler is a tensor $g \in \mathbb{R}^{\hinput \!\times \winput \!\times \htransform \!\times \wtransform \!\times 2}$, $(g[k,l,p,q,0], g[k,l,p,q,1]) :=  G_{kl}(p,q)$ with coordinates w.r.t.\ the input feature map.

\textbf{Recognition scores.}
The next step is to use the grids of matching points to extract scores $s \in \mathbb{R}^{\hinput \!\times \winput \!\times 1}$ indicating how likely the location has a detection.
The soft-inlier count~\cite{rocco2018weakalign} is a natural candidate for this score, however computing it requires enormous amount of the device memory.
Specifically, one needs to create a mask of the size $\hinput \!\times \winput \!\times \htransform \!\times \wtransform \!\times \htransform \!\times \wtransform$\!, which is $\htransform \wtransform \!= 225$ times larger than any tensor we have created so far.
To circumvent the problem, we use a related but different quantity, which is more efficient to compute in a fully-convolutional way.
We directly resample the correlation tensor~$c$ w.r.t.\ the computed grids, i.e., $\hat{s}[k,l,p,q] := c[g[k,l,p,q,0],g[k,l,p,q,1],p,q]$, $\hat{s} \in \mathbb{R}^{\hinput \!\times \winput \!\times \htransform \!\times \wtransform}$\!\!.
The values $c[y,x,p,q]$ at non-integer points $(y, x)$ are computed by differentiable bilinear resampling~\cite{jaderberg2015} on the 2D array $c[:,:,p,q]$.
Note that this operation is not directly supported by the standard bilinear interpolation (different channels need to be resampled at different points).
One can either loop over all the channels and resample them sequentially (can be slow) or create a specialized layer.
The last step to get $s$ is to pool $\hat{s}$ w.r.t.\ its two last dimensions.
We use the average pooling and omit the boundary of the grids to reduce effect of background matches.

\textbf{Localization boxes.}
We extract the localization of the detections by taking $\max$ and $\min$ of grid tensor $g$ w.r.t.\ its 3rd and 4th dimensions, i.e., output the tight bounding boxes around the transformed grid points.

\section{Training the model \label{sec:training}}
\textbf{Training batches and data augmentation.}
In our datasets, input images are of high resolution and contain many objects (see Figure~\ref{fig:results} for examples).
We can't downsample them to a small fixed size (as typical in object detection) because of the strong distortion of the aspect ratio, and each object might simply get too few pixels.
Instead, when constructing a batch we randomly choose a scale and location at which the current image will be processed and resample it to provide a training image of the target size (random crop/scale data augmentation).
For each batch, we collect a set of class images (we cannot use all classes, because there are too many) by taking the annotated classes of the batch and adding some random classes as negatives to fill the class batch size.

\textbf{Objective function.}
As in regular object detection we build the training objective from recognition and localization losses: the hinge-embedding loss with margins for recognition and the smoothed $L_1$ loss for localization:
\begin{align}
\ell_{\irec}^{\ipos}(s) &=  \max( m_{\ipos} - s, 0 ), \;\;\;\;
\ell_{\irec}^{\ineg}(s) =  \max( s - m_{\ineg}, 0 ), \\
\ell_{\iloc}(\vec{x}, \vec{y}) &= \sum_{c=1}^4\left\{
\begin{aligned}
&\tfrac12 ( x_c - y_c)^2, \;\text{if $|x_c-y_c|<1$}, \\
&|x_c-y_c| - \tfrac12, \;\text{otherwise}.
\end{aligned}
\right.
\end{align}
Here $s \in [-1,1]$ is the recognition score (trained to be high for positives and low for negatives), $ m_{\ineg}$ and $m_{\ipos}$ are the negative and positive margins, respectively, $\vec{x}, \vec{y} \in \mathbb{R}^4$ are the output and target encodings of bounding boxes (we use the standard encoding described, e.g., in~\cite[Eq.~2]{ren2015faster}).

As in detection and retrieval, our task has an inherent difficulty of non-balanced number of positives and negatives, so we need to balance summands in the objective.
We started with the Faster R-CNN approach~\cite{ren2015faster}: find a fixed number of hardest negatives per each positive within a batch (positive to negative ratio of 1:3).
The localization loss is computed only for the positive objects.
All the losses are normalized by the number of positives $n_{\ipos}$ in the current batch.
For recognition, we start with the contrastive loss from retrieval~\cite{mobahi2009,simoserra2015,radenovic2018}: squared $\ell_{\mbox{rec}}$ and the positive margin $m_{\mbox{pos}}$ set to $1$ (never active as $s \in [-1,1]$).
\begin{align}
\label{eq:contrastiveloss}
\mathcal{L}_{\irec}^{\mbox{\scriptsize CL}} &=
\tfrac{1}{n_{\ipos}} \sum_{i: t_i=1}\ell_{\irec}^{\ipos}(s_i)^2 + 
\tfrac{1}{n_{\ipos}} \sum_{i: t_i=0}\ell_{\irec}^{\ineg}(s_i)^2,
\\
\label{eq:localizationloss}
\mathcal{L}_{\iloc} &= \tfrac{1}{n_{\ipos}} \sum_{i: t_i=1}\ell_{\iloc}(\vec{x}_i, \vec{y}_i).
\end{align}
Here, the index~$i$ loops over all anchor positions, $\{ s_i, \vec{x}_i\}_i$ come from the network and $\{t_i, \vec{y}_i\}_i$ come from the annotation and assigned targets (see below).

We also tried the recently proposed ranked list loss (RLL) of Wang et~al.~\cite{wang2019rll}, which builds on the ideas of Wu et~al.~\cite{wu2017sampling} to better weight negatives. We have
\begin{align}
\label{eq:rllloss}
\mathcal{L}_{\irec}^{\mbox{\scriptsize RLL}}
&=
\sum_{i: t_{i}=1} \!\tfrac{1}{\tilde{n}_{\ipos}}\ell_{\irec}^{\ipos}(s_{i})
+
\sum_{i: t_{i}=0} \!w^{\ineg}_{i}\ell_{\irec}^{\ineg}(s_{i}),
\\
w^{\ineg}_{i} &\propto \exp( T \ell_{\irec}(s_{i}, 0) )[\ell_{\irec}^{\ineg}(s_{i}) > 0].
\end{align}

Here $\tilde{n}_{\ipos}$ is the number of active positives, i.e., positives such that $\ell_{\irec}^{\ipos}(s_{i}) > 0$.
The weights $w^{\ineg}_{i}$ are normalized in such a way that they sum to 1 over all the negatives for each image-class pair.
The constant~$T$ controls how peaky the weights are.
Wang et~al.~\cite{wang2019rll} fixed~$T$ in advance, but we found it hard to select this parameter.
Instead, we chose it adaptively for each image-class pair in such a way that the weight for the negative with the highest loss is $10^3$ times larger that the weights of the negatives  at the margin boundary, i.e., $s_{i} = m_{\ineg}$.
Finally, we did not back-propagate gradients through the weights~$w^{\ineg}_{i}$ keeping those analogous to probabilities used for sampling negatives.

\textbf{Target assignment.}
For each position of the feature map extracted from the input image~$\iminput$, we assign an anchor location w.r.t.\ which we decode from the output of the transformation net.
At each location, as the anchor we use the rectangle corresponding to the receptive field of the transformation net.
The next step is to assign targets (positive or negative) to all the anchor-class pairs and feed them into the loss functions as targets.
First, we tried the standard object detection strategy, i.e., assign positive targets to the anchors with intersection over union (IoU) with a ground-truth object above 0.5 and negatives~-- to all anchors with IoU $< 0.1$.
Note that we cannot force each ground-truth object to have at least one positive anchor, because we process an image in a training batch at only one scale, which might differ from the scale of the annotated objects.
With the latter constraint, our set of positive anchors is sparser than typical in object detection.
We tried to lower the IoU threshold for positives, but it never helped.
To overcome this problem, we propose to use the standard IoU threshold only to determine localization targets (bounding boxes), and decide for classification targets \emph{after} the output localization is computed.
There, we set the high IoU thresholds of 0.8 for positives and 0.4 for negatives.
We refer to this technique as \emph{target remapping}.
Target remapping is suitable for our model because we compute recognition scores at locations different from the anchors (due to the transformation model), which is not the case in regular object detection models.

\section{Related works \label{sec:relatedworks}}

\textbf{Object detection.}
The standard formulation of object detection assumes a fixed list of target classes (usually 20-80) and an annotated dataset of images (preferably of a large size), where all the objects of the target classes are annotated with bounding boxes.
Modern neural-network detectors are of two types: two-stage and one-stage.
The two-stage detectors follow Faster R-CNN~\cite{girshick2015fast,ren2015faster} and consist of two stages: region proposal (RPN) and detection networks.
RPN generates bounding boxes that might contain objects, and the detection network classifies and refines them; a special pooling operation connects the two nets.
In this approach, the RPN is, in fact, a detector of one joint \emph{object} class.
The one-stage methods like YOLO~\cite{redmon2016,redmon2017}, SSD~\cite{liu2016ssd}, RetinaNet~\cite{lin2017retina} use only RPN-like networks and are often faster but less accurate than  two-stage methods.
Our \methodname architecture is a one-stage method and never computes the detections of a general object, which potentially implies better generalization to unseen classes.

\textbf{One-shot object detection.}
Several recent works~\cite{chen2018lstd,karlinsky2018,michaelis2018,hsieh2019coae,kang2019,wang2019,yan2019,fan2020,perez-rua2020,wang2020} tackled the problems of one-shot and few-shot detection, and \cite{karlinsky2018,hsieh2019coae,fan2020} are most relevant for us.
Karlinsky et~al.~\cite{karlinsky2018} built a two-stage model where RPN did not depend on the target class and recognition was done using the global representation of the target class image.
Hsieh et~al.~\cite{hsieh2019coae} and Fan et~al.~\cite{fan2020} used attention mechanisms to influence the RPN object proposals (thus having non class-agnostic proposals) and later build context dependent global representations.
Our model differs from the works~\cite{karlinsky2018,hsieh2019coae,fan2020} in the two key aspects:
(1) we do not build one global representation of the image of the target classes, but work with lower-level features and match feature map to feature map instead of vector to vector;
(2) our model is one-stage and never outputs non-class specific bounding boxes, which helps to generalize to unseen classes.

\textbf{Image retrieval.}
Image retrieval aims to search a large dataset of images for the ones most relevant to a query image.
A specific example of such a formulation is to search for photos of a building given one photo of the same building.
Differently from one-shot detection, the output is a list of images without object bounding boxes.
Most modern methods for image retrieval model relevance via distances between global image representations obtained by pooling neural network features.
These representations can rely on pre-trained networks~\cite{babenko2014} or, as common recently, be learned directly for image retrieval~\cite{arandjelovic2017netvlad,gordo2017,radenovic2018,wang2019rll}.
Differently, the work of  Noh et~al.~\cite{noh2017} matches local image features similarly to \methodname.
However, differently from \methodname, they do not match densely extracted features directly but subsample them with an attention mechanism, train the network not end-to-end (several steps), and the spatial verification of matches (RANSAC) is not a part of their network but a post-processing step.

\textbf{Tracking in video.} Visual object tracking resembles one-shot detection if we think about the marked object on the first frame as the class image and the remaining video frames as the input images.
Starting from the fully-convolutional Siamese model~\cite{bertinetto2016} matching features with correlation has become a popular approach in tracking.
Recent well-performing systems~\cite{valmadre2017,li2018tracking} introduce asymmetry into the branches before computing the correlation.
\methodname computes correlations with the features extracted by identical networks.
We tried breaking this symmetry similarly to~\cite{li2018tracking}, but it severely degraded performance on the unseen classes.

\textbf{Semantic alignment.}
The works of Rocco et~al.~\cite{rocco2017,rocco2018pami,rocco2018weakalign} studied the problem of semantic alignment, where the goal is to compute correspondence between two images with objects of the same class, and the primary output of their methods (what they evaluate) is the sparse set of matching keypoints.
In one-shot detection, we are interested in the score representing similarity and the bounding box around the object, which is related, but different.
Our model reuses the TransformNet architecture of~\cite{rocco2017,rocco2018pami,rocco2018weakalign} (network that takes dense correlation maps as input and outputs the transformation parameters).
Differently from these works, we apply this network in a fully-convolutional way, which allows us to process input images of arbitrarily large resolution.
As a consequence, we need to resample correlation maps in a fully-convolutional way, which makes the existing network head inapplicable.
Another line of comparison lies in the way of training the models.
Our approach is similar to~\cite{rocco2018weakalign}, where the transformation is trained under weak supervision, whereas Rocco et~al.~\cite{rocco2017,rocco2018pami} used strong supervision coming from synthetically generated transformations.

\section{Experiments \label{sec:experiments}}
\textbf{Datasets and evaluation.} 
Most of prior works~\cite{chen2018lstd,karlinsky2018,michaelis2018,kang2019,wang2019,yan2019,hsieh2019coae} constructed one-shot detection dataset by creating episodes from the standard object detection datasets (PASCAL \cite{everingham10} or COCO\cite{lin2014coco}) where the class image was randomly selected from one of the dataset images.
However, we believe that this approach poses a task that requires either strong assumptions about what the target classes are (inevitably limit generalization to new classes, which is our main goal) or richer definitions of the target class (e.g., more than one shot).
This happens because in the standard object detection the classes are very broad, e.g., contain both objects and their parts annotated as one class, or the objects look very different because of large viewpoint variation.
A large number of training images allows to define such classes, which is not possible in the one-shot setting (e.g., detect a full dog given only a head as the class image or detect the front of a car given its side view).
In other words, the difference of the previous settings to ours is similar to the difference of instance-level to semantic retrieval~\cite{gordo2017semantic}.

Having these points in mind, we chose to evaluate in domains where the one-shot setting is more natural.
Specifically, we used the GroZi-3.2k dataset~\cite{george2014grozi} for retail product detection as the development data.
We create consistent bounding box annotations to train a detection system and collected extra test sets of retail products to evaluate on unseen classes.
Additionally, we evaluate our methods on the INSTRE dataset~\cite{wang2015instre} originally collected for large scale instance retrieval.
The dataset is hard due to occlusions and large variations in scales and rotations~\cite{iscen2017instre}.
The dataset has classes representing physical objects in the lab of the dataset creators and classes collected on-line (buildings, logos and common objects).
We refer to the two parts of the datasets as INSTRE-S1 and INSTRE-S2, respectively.
We provide the details about the dataset preparation in Appendix~\ref{sec:experiments:datasets}.

In all experiments, we use the standard Pascal VOC metric~\cite{everingham10}, which is the mean average precision, mAP, at the intersection-over-union (IoU) threshold of 0.5.
The metric uses the ``difficult'' flag to effectively ignore some detections.

For completeness, we report the results of our methods on the ImageNet-based setup of~\cite{karlinsky2018} in Appendix~\ref{sec:imagenet}.
Confirming the difference of the tasks, \methodname is not competitive but our main baseline (see Section~\ref{sec:exp:baselines}) slightly outperforms the method of~\cite{karlinsky2018}.

\begin{table}[t]
        \caption{Validation mAP of different \methodname training configs.
        The V1-train and V2-train configs selected for experiments in Table~\ref{tbl:init} are marked with ``{\footnotesize$\P$}'' and ``{\footnotesize$\S$}'', respectively\label{tbl:ablation}}
\centering
        \begin{tabular}{@{}l@{\;}|@{\;}l|@{\;}c@{\;\;}c@{\;\;}c@{\;}c@{\;}|@{\;}c@{\;\;}c@{\;\;}c@{\;\;}c@{\;}c@{\;}c@{\;\;}c@{\;\;}c@{}}
            \toprule
            & Config options         &    \multicolumn{12}{c}{Training configs}  \\ %
            \midrule
            & TransformNet          &  \multicolumn{4}{c@{\;}|@{\;}}{V1}   &  \multicolumn{8}{c}{V2} \\
            \midrule
            \multirow{5}{*}{\rotatebox{90}{\centering training}} 
            & loss                   &  {\scriptsize CL}  & {\scriptsize RLL}  & {\scriptsize RLL}  &  {\scriptsize RLL} &  {\scriptsize RLL}  & {\scriptsize RLL}  & {\scriptsize RLL}  & {\scriptsize RLL}  & {\scriptsize RLL}  & {\scriptsize CL} & {\scriptsize CL} & {\scriptsize RLL}  \\
            & remap targets          &      &      &  +   &  +   &  +    & +    & +    & +    & +    &    & +  &   \\
            & mine patches           &      &      &      &  +   &       & +    &      &      & +    &    &    &   \\
            & init transform         &      &      &      &      &       &      & +    & +    & +    & +  & +  & + \\
            & zero loc loss          &      &      &      &      &       &      &      & +    & +    & +  & +  & + \\
            \midrule
            & mAP                    & 81.8 & 85.7 & 87.0$^\P$ & 86.5 & 87.0  & 86.5 & 88.0 & 89.5$^\S$ & \textbf{90.6} & 87.4 & 88.1 & 87.1 \\
            \bottomrule
        \end{tabular}
\end{table}

\subsection{Ablation study \label{sec:exp:detretnet}}

\textbf{Model architecture.}
The \methodname model contains two subnets with trainable parameters: TransformNet and feature extractor.
For TransformNet, we use the network of the same architecture as Rocco et~al.~\cite{rocco2017,rocco2018weakalign} and plug it into our model in the two ways: (V1) simplified affine transformations with $P = 4$ (only translation and scaling) applied in the direction natural to \methodname; (V2) full affine transformations with $P = 6$ applied in another direction (used in~\cite{rocco2017,rocco2018weakalign}).
The V1 approach allows to have full supervision (the annotated bounding boxes define targets for the 4 parameters of the simplified transformations) and is more convenient to use (no need to invert transformations).
The V2 approach as in~\cite{rocco2018weakalign} requires a form of weak supervision for training, but can be initialized from the weights released by~Rocco et~al.~\cite{rocco2017,rocco2018weakalign}, and, surprisingly is fully-functional without any training on our data at all.
Comparison of V1 vs. V2 in different settings is shown in Tables~\ref{tbl:ablation}, \ref{tbl:init}, \ref{tbl:baselines}, \ref{tbl:results_instre} (see full descriptions below). 
For the features extractor, we consider the third block of ResNet-50 and ResNet-101.
See Table~\ref{tbl:init} for the comparison (the description is below).

\textbf{Training configurations.}
We now evaluate multiple options of our \methodname training config (see Table~\ref{tbl:ablation} for the results).
We experiment with the ResNet-50 feature extractor initialized from the weights trained on ImageNet~\cite{ILSVRC15} for image classification.
For each run, we choose the best model by looking at the mAP on the validation set \texttt{val-new-cl} over the training iterations and report this quantity as mAP.

We start with the V1 approach and training with the standard contrastive loss for recognition~\eqref{eq:contrastiveloss} and smoothed L1 loss for localization~\eqref{eq:localizationloss} and gradually add more options.
We observed that RLL~\eqref{eq:rllloss} outperformed the contrastive loss \eqref{eq:contrastiveloss}, recognition target remapping (see Section~\ref{sec:training}) helped.
In addition to the performance difference, these feature regularized the training process (the initial models were quickly starting to overfit).
Finally, we implemented a computationally expensive feature of mining hard patches to feed into batches (like hard negative mining of images for image retrieval~\cite{arandjelovic2017netvlad,radenovic2018}).
This step significantly slowed down training but did not improve V1 models.%

Next, we switch to the V2 approach.
The main benefit of V2 is to initialize TransformNet from the weights of~Rocco et~al.~\cite{rocco2017,rocco2018weakalign}, but training runs without target remapping worsen the mAP at initialization.
We also found that training signals coming to TransformNet from the localization and recognition losses were somewhat contradicting, which destabilized training.
We were not able to balance the two, but found that simply zeroing out the localization loss helped (see Figure~\ref{fig:model} bottom-left for a visualization of the learned transformations).
Finally, we re-evaluated the options, which helped V1, in the context of V2 and confirmed that they were important for successful training.
Mining hard patches now improved results (see Table~\ref{tbl:ablation}).

Finally, we selected the best config for V1 as V1-train and the best config overall as V2-train (we use V2-init to refer to its initialization).
The additional implementation details are presented in Appendix~\ref{sec:impldetails}.

We also separately evaluated the effect of using the inverse and full affine transformations~-- they did not hurt mAP, but when visualizing the results we did not see the model learning transformations richer than translation and scaling, so we omitted the architectures in-between V1 and V2.

\begin{table}[t]
        \caption{Initializations for the \methodname feature extractor  (mAP on the validation set). Symbol ``$\dagger$'' marks the ImageNet models converted from Caffe, symbol ``$\ddagger$'' marks the model with group norm~\cite{wu2018groupnorm} instead of batch norm\label{tbl:init}\vspace{-2mm}}
    \centering
        \begin{tabular}{@{}l@{\;\;\;}ll@{\;\;}r@{\;\;}r@{\;\;}r@{}}
            \toprule
            & Src task & Src data & V1-train &  V2-init & V2-train \\
            \midrule
            \multirow{ 5}{*}{\rotatebox{90}{\centering\footnotesize ResNet-50}}
            & ---                               & ---                                      & 59.6  &     2.0 &     67.9   \\
            & {\footnotesize classification }   & {\footnotesize ImageNet   }              & 84.8  &    79.6 &     89.0   \\
            & {\footnotesize classification$\dagger$ }   & {\footnotesize ImageNet   }     &  87.1 &    \textbf{86.1} &     \textbf{89.5}   \\
            & {\footnotesize classification$\dagger\ddagger$ }   & {\footnotesize ImageNet}&  83.2 &    68.2 &     87.1   \\
            & {\footnotesize detection      }   & {\footnotesize COCO       }              &  81.9 &    77.5 &     88.1   \\
            \midrule
            \multirow{5}{*}{\rotatebox{90}{\centering\footnotesize ResNet-101}}
            & {\footnotesize classification }   & {\footnotesize ImageNet   }              &  85.6 &    81.1 &     89.4  \\
            & {\footnotesize classification$\dagger$ }   & {\footnotesize ImageNet }       & 87.5  &    81.0 &     88.8  \\
            & {\footnotesize retrieval      }   & {\footnotesize landmarks   }             &  \textbf{88.7} &    83.3 &     89.0  \\
            & {\footnotesize detection      }   & {\footnotesize COCO        }             & 85.6  &    73.1 &     86.8 \\
            & {\footnotesize alignment      }   & {\footnotesize PASCAL      }             & 85.7  &    77.2 &     88.7 \\
            \bottomrule
        \end{tabular}
    \vspace{-5mm}
\end{table}

\textbf{Initialization of the feature extractor.}
We observed that the size of the GroZi-3.2k dataset was not sufficient to train networks from scratch, so choosing the initialization for the feature extractor was important.
In Table~\ref{tbl:init}, we compared initializations from different pre-trained nets.
The best performance was achieved when starting from networks trained on ImageNet~\cite{ILSVRC15} for image classification and the Google landmarks dataset~\cite{noh2017} for image retrieval.
Surprisingly, detection initializations did not work well, possibly due to the models largely ignoring color.
Another surprising finding is that the V2-init models worked reasonably well even without any training for our task.
The pre-trained weights of the affine transformation model~\cite{rocco2018weakalign} appeared to be compatible not only with the feature extractor trained with them, likely due to the correlation tensors that abstract away from the specific features.

\textbf{Running time.} 
The running time of the feature extractor on the input image depends on the network size and is proportional to the input image size.
The running time of the feature extractor on the class images can be shared across the whole dataset and is negligible.
The running time of the network heads is in proportional to both the input image size and the number of classes to detect, thus in the case of a large number of classes dominates the process.
On GTX 1080Ti in our evaluation regime (with image pyramid) of the \texttt{val-new-cl} subset, our PyTorch~\cite{paszke2019pytorch} code computed input features in 0.46s per image and the heads took 0.052s and 0.064s per image per class for V1 and V2, respectively, out of which the transformation net itself took 0.020s.
At training, we chose the number of classes such that the time on the heads matched the time of feature extraction.
Training on a batch of 4 patches of size 600x600 and 15 classes took~0.7s.

\subsection{Evaluation of \methodname against baselines \label{sec:exp:baselines}}

\textbf{Class detectors.}
We started with training regular detectors on the GroZi-3.2k dataset.
We used the maskrcnn-benchmark system~\cite{massa2018mrcnn} to train  the Faster R-CNN model with Resnet-50 and Resnet-101 backbones and the feature pyramid network to deal with multiple scales~\cite{lin2017fpn}.
However, these systems can detect only the training classes, which is the \texttt{val-old-cl} subset.

\textbf{Main baseline: object detector + retrieval.}
The natural baseline for one-shot detection consists in combining a regular detector of all objects merged into one class with the image retrieval system, which uses class objects as queries and the detections of the object detector as the database to search for relevant images.
If both the detector and retrieval are perfect then this system solves the problem of one-shot detection.
We trained object detectors with exactly same architectures as the class detectors above.
The ResNet-50 and ResNet-101 versions (single class detection) delivered on the validation images (combined \texttt{val-old-cl} and \texttt{val-new-cl}) 96.42 and 96.36 mAP, respectively.
For the retrieval, we used the software\footnote{\url{https://github.com/filipradenovic/cnnimageretrieval-pytorch}} of~Radenovi\'{c} et~al.~\cite{radenovic2016,radenovic2018}.
We trained the models on the training subset of GroZi-3.2k and chose hyperparameters to maximize mAP on the \texttt{val-new-cl} subset.
Specifically, GeM pooling (both on top of ResNet-50 and ResNet-101) and end-to-end trainable whitening worked best (see Appendix~\ref{sec:baselinedetails} for details).

\textbf{Sliding window baselines.}
To evaluate the impact of our transformation model, we use the same feature extractor as in \methodname (paired with the same image pyramid to detect at multiple scales), but omit the transformation model and match the feature map of the class image directly with the feature map of the input image in the convolutional way.
In the first version (denoted as ``square''), we used the feature maps resized to squares identical to the input of the transformation model (equivalent to fixing the output of the transformation model to identity).
In the second version (denoted as ``target AR''), we did not resize the features and used them directly from the feature extractor, which kept the aspect ratio of the class image correct.

\textbf{Extra baselines.}
We have compared \methodname against the recently released code\footnote{\url{https://github.com/timy90022/One-Shot-Object-Detection}} of Hsieh et~al.~\cite{hsieh2019coae} (see Appendix~\ref{sec:baselinedetails} for details).
We also compared against other available codes~\cite{michaelis2018,karlinsky2018}  with and without retraining the systems on our data but were not able to obtain more than 40~mAP on \texttt{val-new-cl}, which is very low, so we did not include these methods in the tables.

\begin{table}[t]
        \caption{Comparison to baselines, mAP, on the task of retail product detection. *In this version of the main baseline, the detector is still trained on the training set of GroZi-3.2k, but the retrieval system uses the weights trained on ImageNet for classification\label{tbl:baselines}\vspace{-2mm}}
\centering
        \begin{tabular}{@{}l@{\!\!\!\!}r@{\;\;}r@{\;\;}r@{\;\;}r@{\;\;}r@{\;\;}r@{}}
            \toprule
            Method     & Trained & \texttt{val-old-cl} & \texttt{val-new-cl} & \texttt{dairy} & \texttt{paste-v} & \texttt{paste-f} \\
            \midrule
            Class detector & yes &  87.1 & --- & --- & --- & --- \\
            \midrule
            Main baseline  & no*\!\!\!      & 72.0 & 69.1 & 46.6 & 34.8 & 31.2 \\
            Main baseline  & yes            & 87.6 & 86.8 & 70.0 & 44.3 & 40.0 \\
            \midrule
            Sliding window, square    &  no        &  57.6 & 58.8 & 33.9 & 8.0  & 7.0  \\
            Sliding window, target AR &  no        &  72.5 & 71.3 & 63.0 & 65.1 & 45.9 \\
            \midrule
            Hsieh et~al.~\cite{hsieh2019coae} &  yes       &  \textbf{91.1} & 88.2 & 57.2 & 32.6 & 27.6 \\
            \midrule
            ours: \methodname V1-train & yes   & 89.1 & 88.7 & 70.5 & 61.9 & 48.8 \\
            ours: \methodname V2-init  & no    & 79.7 & 86.1 & 65.4 & 68.2 & 48.4 \\
            ours: \methodname V2-train & yes   & 85.0 & \textbf{90.6} & \textbf{71.8} & \textbf{73.3} & \textbf{54.5} \\
            \bottomrule
        \end{tabular}
\end{table}

\begin{table}[t]
    \caption{Results on the INSTRE dataset, mAP  \label{tbl:results_instre}}
\centering
\vspace{-2mm}
        \begin{tabular}{@{}lr@{\;\;\;\;}r@{\;\;\;\;}r@{\;\;\;\;}r@{}}
            \toprule
            & \multicolumn{2}{c}{ResNet-50} & \multicolumn{2}{c}{ResNet-101} \\
            \midrule
            Method & INSTRE-S1 & INSTRE-S2 & INSTRE-S1 & INSTRE-S2\\
            \midrule
            Main baseline-train      & 72.2 & 64.4 & 79.0 & 53.9 \\
            Sliding window, AR       & 64.9 & 57.6 & 60.0 & 51.3 \\
            Hsieh et~al.~\cite{hsieh2019coae}    & 73.2 & 66.7 & 68.1 & 74.8 \\
            ours: \methodname V1-train    & 83.9 &      73.8 & 87.1 &      76.0 \\
            ours: \methodname V2-init     & 71.9 &      64.5 & 69.7 &      63.2 \\
            ours: \methodname V2-train    & \textbf{88.7} &      \textbf{77.7} & \textbf{88.7} &      \textbf{79.5} \\
            \bottomrule
        \end{tabular}
    \vspace{-5mm}
\end{table}

\textbf{Comparison with baselines.}
Tables~\ref{tbl:baselines} and~\ref{tbl:results_instre} show quantitative comparison of our \methodname models to the baselines on the datasets of retail products and the INSTRE datasets, respectively.
Figure~\ref{fig:results} shows qualitative comparison to the main baseline (see Appendix~\ref{sec:extraresults} for more qualitative results).
Note, that the datasets \texttt{paste-f}, INSTRE-S1 and INSTRE-S2 contain objects of all orientations.
Features extracted by CNNs are not rotation invariant, so as is they do not match.
To give matching-based methods a chance to work, we augment the class images with their 3 rotations (90, 180 and 270 degrees) for all the methods.

First, note that when evaluated on the classes used for training some one-shot method outperformed the corresponding class detectors, which might be explained by having too little data to train regular detectors well enough.
Second, in all the cases of classes unseen at the training time, the trained \methodname versions outperform the baselines.
Qualitatively, there are at least two reasons for such significant difference w.r.t.\ our main baseline:
the object detector of the baseline has to detect objects without knowing what class is has to detect, which implies that it is very easy to confuse object with its part (see examples in Figure~\ref{fig:teaser} and Figure~\ref{fig:results}-right) or merge multiple objects together;
the current retrieval system is not explicitly taking the geometry of matches into account, thus it is hard to distinguish different objects that consist of similar parts.
Finally, note that the sliding window baseline with the correct aspect ratio performed surprisingly well and sometimes outperformed the main baseline, but the learned transformation model always brought improvements.

One particular difficult case for the matching-based \methodname models appears to be rotating in 3D objects (see wrong detection in Figure~\ref{fig:results}-mid).
The model matches the class image to only half of the object, which results in producing incorrect detection localization.
Richer and maybe learnable way of producing bounding boxes from matched features is a promising direction for future work.

\begin{figure}[t]
    \begin{center}
        \begin{tabular}[t]{@{}p{0.33\textwidth}@{}p{0.33\textwidth}@{}p{0.33\textwidth}@{}}
            \vtop{\vskip 0pt \vskip -\ht\strutbox 
                \begin{minipage}{0.33\textwidth}
                    \begin{tabular}[t]{@{}p{0.05\textwidth}@{\;}p{0.95\textwidth}@{}}
                        \rotatebox{90}{\centering\footnotesize classes} &
                        \begin{overpic}[width=0.30\textwidth]{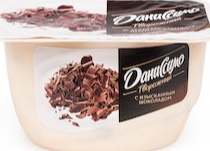}
                            \put (0,0){\colorbox{white}{\scriptsize 11}}
                        \end{overpic}
                        \begin{overpic}[width=0.25\textwidth]{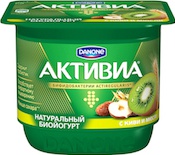}
                            \put (0,0){\colorbox{white}{\scriptsize 22}}
                        \end{overpic}
                        \begin{overpic}[width=0.10\textwidth]{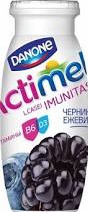}
                            \put (0,0){\colorbox{white}{\scriptsize 25}}
                        \end{overpic}
                        \begin{overpic}[width=0.10\textwidth]{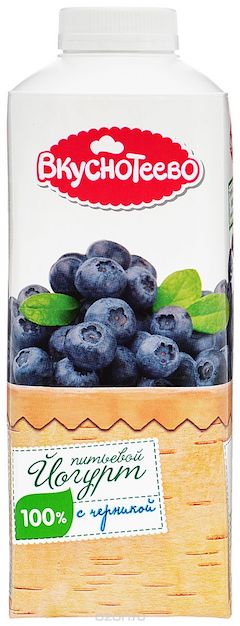}
                            \put (0,0){\colorbox{white}{\scriptsize 132}}
                        \end{overpic}
                        \\
                        \rotatebox{90}{\quad\footnotesize baseline} &
                        \includegraphics[width=0.9\textwidth, clip=true, trim=10mm 12mm 0mm 17.2mm]{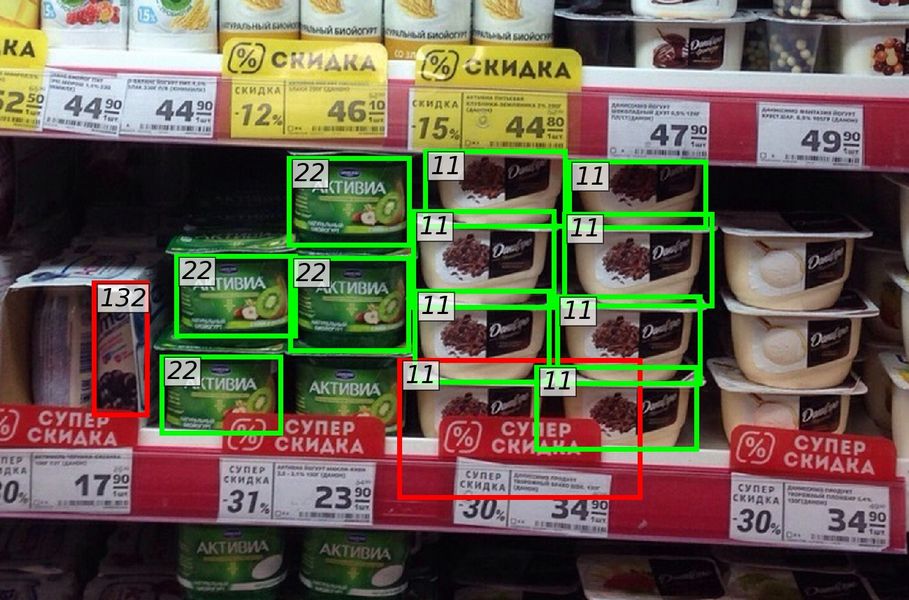}
                        \\[-1mm]
                        \rotatebox{90}{\;\;\quad\footnotesize \methodname} &
                        \includegraphics[width=0.9\textwidth, clip=true, trim=10mm 12mm 0mm 17.2mm]{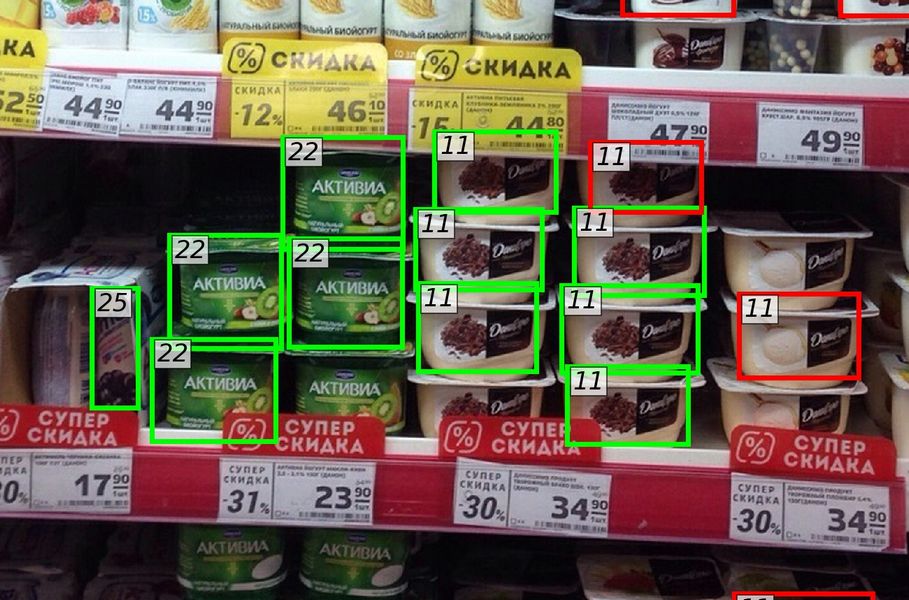}
                    \end{tabular}
                \end{minipage}
                \vskip -\dp\strutbox }%
            &
            \vtop{\vskip 0pt \vskip -\ht\strutbox 
                \begin{minipage}{0.33\textwidth}
                    \begin{tabular}[t]{@{}p{0.05\textwidth}@{\;}p{0.95\textwidth}@{}}
                        \rotatebox{90}{\centering\footnotesize classes} &
                        \begin{overpic}[width=0.205\textwidth]{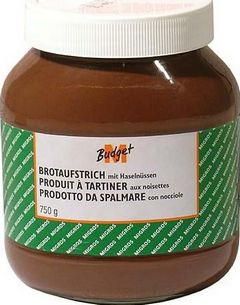}
                            \put (0,0){\colorbox{white}{\scriptsize 150}}
                        \end{overpic}
                        \begin{overpic}[width=0.205\textwidth]{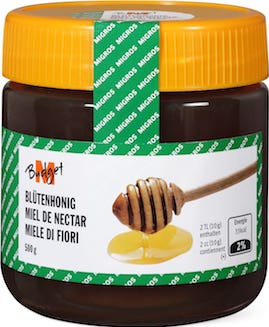}
                            \put (0,0){\colorbox{white}{\scriptsize 151}}
                        \end{overpic}
                        \\
                        \rotatebox{90}{\quad\footnotesize baseline} &
                        \includegraphics[width=0.9\textwidth, clip=true, trim=0mm 0mm 0mm 95mm]{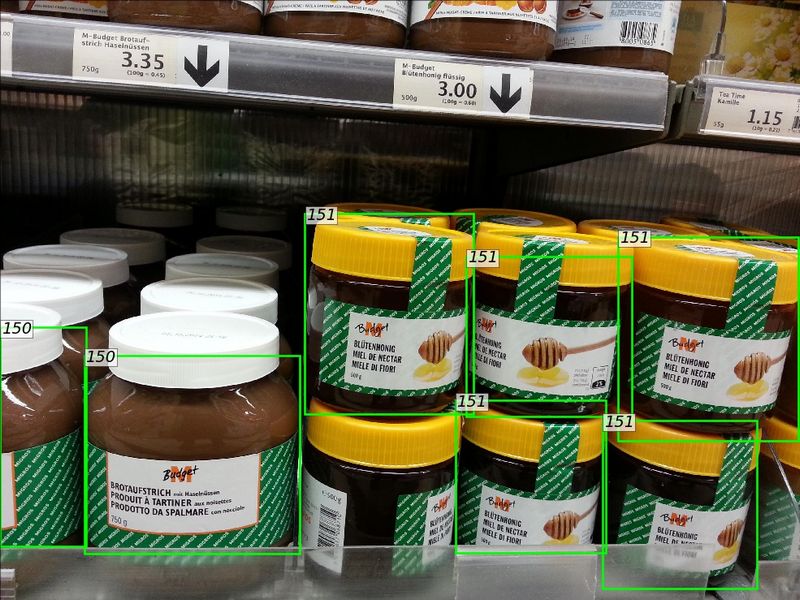}
                        \\[-1mm]
                        \rotatebox{90}{\;\;\quad\footnotesize \methodname} &
                        \includegraphics[width=0.9\textwidth, clip=true, trim=0mm 0mm 0mm 95mm]{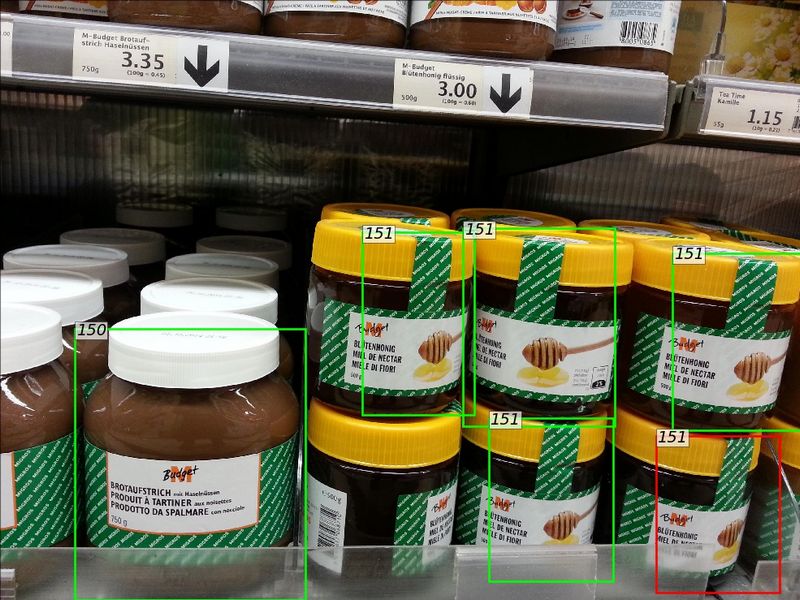}
                    \end{tabular}
                \end{minipage}
                \vskip -\dp\strutbox }%
            &
            \vtop{\vskip 0pt \vskip -\ht\strutbox 
                \begin{minipage}{0.33\textwidth}
                    \begin{tabular}[t]{@{}p{0.05\textwidth}@{\;}p{0.95\textwidth}@{}}
                        \rotatebox{90}{\centering\footnotesize classes} &
                        \begin{tabular}{@{}p{0.25\textwidth}@{}p{0.33\textwidth}@{}p{0.33\textwidth}@{}}
                            \vtop{\vskip 0pt \vskip -\ht\strutbox 
                                \multirow{2}{*}{
                                    \begin{overpic}[height=8mm]{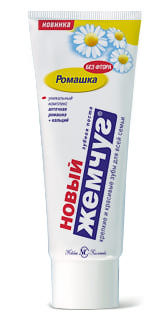}
                                        \put (0,80){\colorbox{white}{\scriptsize 15}}
                                    \end{overpic}
                                    \begin{overpic}[height=8mm]{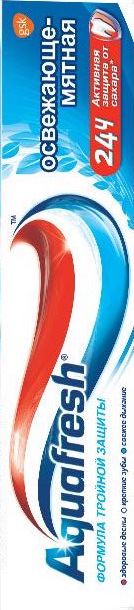}
                                        \put (-5,80){\colorbox{white}{\scriptsize\!\!\!14}}
                                    \end{overpic}
                                }
                                \vskip -\dp\strutbox }%
                            &
                            \vtop{\vskip 0pt \vskip -\ht\strutbox 
                                \begin{overpic}[width=13mm]{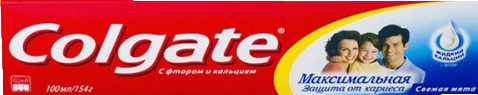}
                                    \put (0,0){\colorbox{white}{\scriptsize 47}}
                                \end{overpic}
                                \vskip -\dp\strutbox }%
                            &
                            \vtop{\vskip 0pt \vskip -\ht\strutbox 
                                \begin{overpic}[width=13mm]{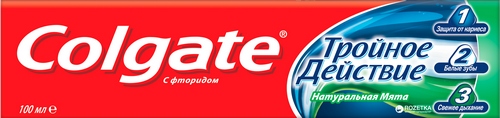}
                                    \put (0,0){\colorbox{white}{\scriptsize 72}}
                                \end{overpic}
                                \vskip -\dp\strutbox }%
                            \\[-3mm]
                            &
                            \vtop{\vskip 0pt \vskip -\ht\strutbox 
                                \begin{overpic}[width=13mm]{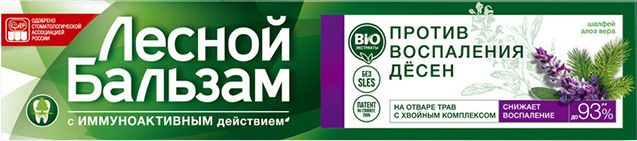}
                                    \put (0,0){\colorbox{white}{\scriptsize 57}}
                                \end{overpic}
                                \vskip -\dp\strutbox }%
                            &
                            \vtop{\vskip 0pt \vskip -\ht\strutbox 
                                \begin{overpic}[width=13mm]{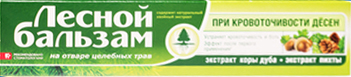}
                                    \put (0,0){\colorbox{white}{\scriptsize 63}}
                                \end{overpic}
                                \vskip -\dp\strutbox }%
                            \\
                        \end{tabular}
                        \\[-3.1mm]
                        \rotatebox{90}{\quad\footnotesize baseline} &
                        \includegraphics[width=0.9\textwidth, clip=true, trim=0mm 8mm 0mm 7.2mm]{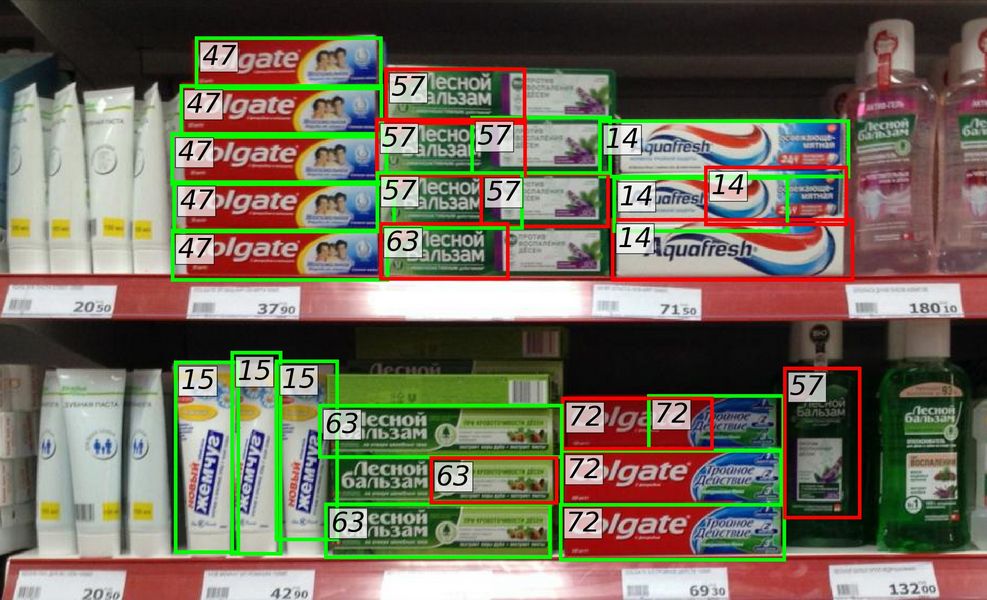}
                        \\[-1.0mm]
                        \rotatebox{90}{\;\;\quad\footnotesize \methodname} &
                        \includegraphics[width=0.9\textwidth, clip=true, trim=0mm 8mm 0mm 7.2mm]{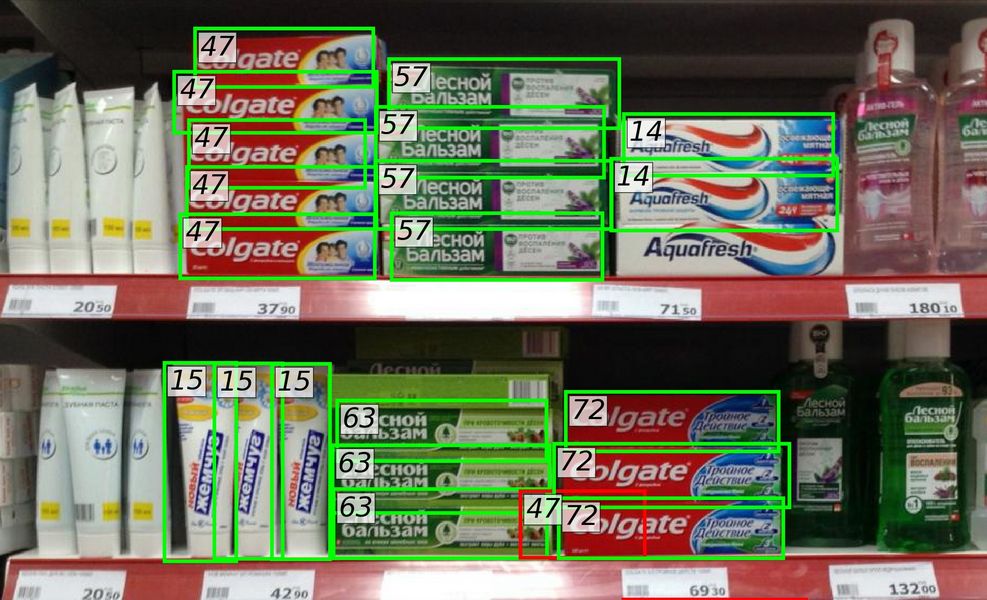}
                    \end{tabular}
                \end{minipage}
                \vskip -\dp\strutbox }%
            \\[-3mm]
            \qquad\texttt{dairy} dataset
            &
            \qquad\texttt{val-new-cl} dataset
            &
            \qquad\texttt{paste-f} dataset\\[-5mm]
        \end{tabular}
    \end{center}
    \caption{Qualitative comparison of the best \methodname model vs.\ the best baseline. Correct detections~-- green boxes, incorrect~-- red boxes\label{fig:results}\vspace{-2mm}}
\end{figure}

\section{Conclusion \label{sec:conclusion}}

In this paper, we proposed the \methodname model for one-shot object detection.
The model combines a deep feature extractor, correlation matching, feed-forward alignment network and bilinear interpolation in a differentiable way that allows end-to-end training.
We trained our model with an objective function combining the recognition and localization losses.
We applied our model to the challenging task of retail product recognition and construct a large dataset with consistent annotation for a large number of classes available (the recent SKU110k dataset~\cite{goldman2019sku} is of larger scale, but contains only object bounding boxes without the class labels).
The \methodname model outperformed several strong baselines, which indicates the potential of the approach for practical usage.

\textbf{Acknowledgments.}
We would like to personally thank Ignacio Rocco, Relja Arandjelovi\'{c}, Andrei Bursuc, Irina Saparina and Ekaterina Glazkova for amazing discussions and insightful comments, without which this project would not be possible.
This research was partly supported by Samsung Research, Samsung Electronics, by the Russian Science Foundation grant 19-71-00082 and through computational resources of HPC facilities at NRU HSE.

\bibliographystyle{splncs04}
\bibliography{references_CR}

\clearpage
\appendix

\vskip .375in
\begin{center}
    {\Large \bf Supplementary Material (Appendix) \par}
    {\Large \bf OS2D: One-Stage One-Shot Object Detection by Matching Anchor Features \par}
    \vspace*{10pt}
\end{center}

\section{Data and evaluation \label{sec:experiments:datasets}}

\subsection{Retail products}
\textbf{Data.}
We used the GroZi-3.2k dataset~\cite{george2014grozi} for retail product detection (680 images collected from 5 different stores) as development data.
We could not use the original annotation of this dataset because it was often grouping multiple similar objects into a single bounding box and had no strict policy about what objects are of the same class and what are of different classes.
Thus, we created new annotation where each object has an individual bounding box, and only identical objects belong to the same class.
For each class, we selected a template class image either from the original annotation (if available) or from the internet by querying the product names.
Importantly, each object was assigned the ``difficult'' flag when the human annotator could not assign a class without guessing.
We ended up having 8921 objects of 1063 classes annotated.

\textbf{Data splits.}
We created the splits of the dataset by selecting 185 classes and keeping the images with those classes away from training: 622 objects in 84 images.
We will refer to this set as \texttt{val-new-cl}.
The selected images also contained 518 objects of classes that appeared at training.
We will refer to this set as \texttt{val-old-cl}.
Throughout our experiments we used \texttt{val-new-cl} as our main validation set, and \texttt{val-old-cl} as a secondary validation showing performance on the training classes.

\textbf{Extra test sets.} To assess the generalization ability of our method, we collected extra test sets containing objects of new classes in the images taken in different conditions.
The \texttt{dairy} set consists of 12 images with 786 objects of 166 classes of dairy products.
The \texttt{paste-f} set consists of 91 images with 4861 objects of 259 classes of toothpaste and accompanying products.
However, the \texttt{paste-f} set contains objects of all orientations, which is different from the training conditions.
We also selected a subset \texttt{paste-v}, where all the objects with incorrect orientation are masked out with the ``difficult'' flag.

Table~\ref{tbl:baselines} of the main paper contains results of the \methodname methods and baselines on the test and validation subsets.

\subsection{Additional dataset: INSTRE}
In addition to the domain of retail products, we applied our methods to the INSTRE dataset~\cite{wang2015instre}, which was originally collected for large scale instance retrieval and has bounding box annotation for all objects.
The dataset is considered hard due to occlusions and large variations in scales and rotations~\cite{iscen2017instre}.
The dataset has 28,543 images and 200 object classes: 100 classes representing physical objects in the lab of the dataset creators, 100 classes collected on-line representing buildings, logos and common objects.
We refer to the classes of the first and second types as INSTRE-S1 and INSTRE-S2, respectively.

The INSTRE dataset is used for evaluating retrieval systems, so does not have splits into train and test.
We modify the evaluation protocol of Iscen et~al.~\cite{iscen2017instre} who selected 5 images of each class as queries (correspond to class images in our terminology) by splitting the classes (with the corresponding images) of both INSTRE-S1 and INSTRE-S2 in the following proportion:
75\% for training, 5\% for validation and 20\% for testing.
In this setting, we can train and evaluate our method and the baselines.
Table~\ref{tbl:results_instre} of the main paper provides results on the two versions on the dataset.
We report the results on the test sets of INSTRE-S1 and INSTRE-S2 after training of the training subsets of INSTRE-S1 and INSTRE-S2, respectively.

\subsection{Evaluation metric.}
In our experiments, we've used the standard Pascal VOC metric~\cite{everingham10}, which is the mean average precision at the intersection-over-union (IoU) threshold of 0.5 (we will refer to it as mAP).
Importantly, this metric also supports the ``difficult'' flag of the annotation:  it is used to exclude ground-truth objects and their detections when computing both recall and precision, which means that the method is neither penalized nor rewarded for detecting objects with this flag.

\section{Implementation details \label{sec:impldetails}}

\subsection{TransformNet architecture \label{sec:transformnet_arch}}
We follow Rocco et~al.~\cite{rocco2017,rocco2018pami,rocco2018weakalign} and use the same architecture of TransformNet: ReLU, channelwise L2-normalization,
conv2d with the kernel $7 \times 7 \times 225 \times 128$, batch norm, ReLU, conv2d with the kernel $5 \times 5 \times 128 \times 64$, batch norm, ReLU, conv2d with the kernel~$5 \times 5 \times 64 \times P$.
Here $P$ is the number of parameters of the transformation, which equals $6$ for the affine transformation.
This network was designed for aligning two feature maps of size $15 \times 15$, i.e., $\htransform = \wtransform = 15$  (corresponds to the image size $240 \times 240$ if using the features after the fourth block of ResNet).

Note that the network starts with ReLU, which corresponds to taking only positive correlations when building transformations (Rocco et~al.~\cite{rocco2017} did not include this layer into TransformNet but applied it right after computing the correlations).

\subsection{OS2D details \label{sec:impldetails:os2d}}

\textbf{Implementation and hardware.}
We implemented the \methodname model based on the PyTorch library~\cite{paszke2019pytorch}.
The models were both trained and tested on GPUs.
The hyperparameters for training were selected to fit the process on Nvidia~GTX~1080~Ti.
However evaluation of retail test sets required more device memory because of higher resolution, small objects and a large quantity of classes.
We used Nvidia~V100 devices for such runs.

\textbf{Training.} The \methodname models were trained with the SGD optimizer for 200k steps with the learning rate of $10^{-4}$, weight decay of $10^{-4}$ and momentum of 0.9.
We decreased the learning rate by a factor of 10 after 100k and 150k training iterations.
We used the input image batch size of 4, cropped patches of size $600 \times 600$ and used at most 15 different labels per batch.
Note that cropping patches of the correct size is effectively a version of random crop/scale data augmentation.
We tried using more types of data augmentation, but none of them was effective.

When training all the models, we converted the switched layers of the feature extractor to the evaluation mode, i.e., did not estimate batch mean and variance.
Keeping batchnorm in the training mode significantly degraded the performance.
When training the V1 and V2 models we kept batchnorm of the transformation network in the training and evaluation modes, respectively.

We followed Rocco et~al.~\cite{rocco2017,rocco2018pami,rocco2018weakalign} and trained TransformNet on only positive pairs.
Technically we achieved this by computing two versions of the transformations at training~-- one with the full computational graph, another with the TransformNet parameters detached from the graph.
The first version was used to train on positives, the second one~-- to train on negatives.
We used this approach because when training transformations on negatives the networks started to ruin the transformation model by moving the transformation in random directions.
On top of that, we often have very similar classes, and we still want them to be aligned properly to better compare the matched features.

When training all V1 models we initialized TransformNet to always output identity transformation by setting the weights of the last convolutional layer of TransformNet to 0 and biases to 0 or 1.

For the objective function, we use the margins $m_{\ipos} = 0.6$ and $m_{\ineg} = 0.5$ when the recognition scores were normalized to the segment $[-1, 1]$.
To train the V1 models, we used the weight of the localization loss of 0.2.
To fine-tune the V2 models, we turned the localization loss completely, i.e., set its weight to zero.

\textbf{Detection.}
Before computing the final results, for all the methods we used the standard non-maximum suppression (NMS) with the IoU threshold of 0.3.
Differently from the maskrcnn-benchmark~\cite{massa2018mrcnn}, we did not do joint NMS for all the classes~-- it always degraded the performance.

At evaluation, we resized the class images with preserving their aspect ratio to have their product of dimensions equal to $240^2$.
For the input images, we used the image pyramid to deal with objects of different scales.
We always use the pyramid of 7 levels: 0.5, 0.625, 0.8, 1, 1.2, 1.4, 1.6 times the dataset scale.
For each dataset, we estimated its scale by computing and rounding the average object size.
The GroZi-3.2k dataset was of scale 1280, the \texttt{dairy} dataset was of scale 3500, the \texttt{paste-v} dataset were also of scale 3500.
However,  the \texttt{paste-v} dataset had too many labels, so the largest image size did not fit into the GPU memory, thus we reduced its scale to 2000 for all experiments with \methodname (the baselines were still run on the initial scale).
Evaluation of an image at a particular scale, e.g., 0.5 * 3500 = 1750, means that we resize the input image such that its largest size equals the scale, e.g., 1750, before feeding it into the feature extractor (or objects detector for the baselines).

\section{Details of the baselines \label{sec:baselinedetails}}
In this section, we describe the details of the baselines that were important to improve their performance.
Note that implementations of both baselines use open-source code, and we provide all the changes and launching scripts together with the OS2D code.\footnote{\url{https://github.com/aosokin/os2d}}

\subsection{Main baseline: detector + retrieval}
For the detector, we used the maskrcnn-benchmark system \cite{massa2018mrcnn}.
We used the Faster R-CNN detectors~\cite{ren2015faster} with the feature pyramid backbone~\cite{lin2017fpn} based on ResNet-50 and ResNet-101.
We used the standard hyperparameters, but added multi-scale training and testing (supported by the library), which were improving results.
The scales of images for both training and testing were set the same to the OS2D training regime.

For the retrieval system used on top of the detections, we used the open-source library by Radenovi\'{c} et~al.~\cite{radenovic2016,radenovic2018}.
We used the trainable Generalized-Mean (GeM) pooling and end-to-end trainable whitening layers.
For the training dataset, we used the class images as queries, annotated detections of the correct/incorrect classes as positives and negatives, respectively.
We also randomly sampled 10 bounding boxes per training image and automatically labeled them as positive/negatives based on their IoU with annotated objects.
In the training process, we used the standard setting with the contrastive loss, hard negative mining, Adam optimizer and learning rate schedule with an exponential decay.
We resized all images (queries, positives and negatives) to have the maximal side equal to 240 (with preserving the aspect ratio).

At the testing stage, we used the same image pyramid as in OS2D for the detector and the multi-scale descriptor (3 scales) for retrieval.

\subsection{CoAE one-shot detector}
We compared our methods with the official implementation of the recent CoAE method of Hsieh et~al.~\cite{hsieh2019coae}.
Their released models (trained on ImageNet) did not generalize well to our settings, so we reported only the results of the retrained models.
For fair comparison with OS2D, we added multi-scale training and testing to the original code.
Multi-scale training helped significantly, while multi-scale testing did not help at all.
In training, we used the same number of iterations as for OS2D (the process converged well) and the same learning rate schedule (but different initial value).

\section{Evaluation in the ImageNet-based setup \label{sec:imagenet}}
In this section, we evaluate our methods on the setting proposed by Karlinsky et~al.~\cite{karlinsky2018}.
The test set is based on the images from 214 categories of the ImageNet-LOC dataset~\cite{ILSVRC15} and is organized in 500 episodes each containing 5 classes (5-way).
Each class of an episode is represented by 10 random images with the class instances.
In the 1-shot setting, each class additionally has one representative: an image with a bounding box around the class instance.
The quality on each episode is measured by the average precision computed on the jointly sorted list of detections of different classes (positive/negative labels are assigned based on the IoU threshold 0.5).
The overall quality is computed as the average AP over episodes.
To distinguish this metric from mAP used in the rest of this paper (where AP is computed for each class independently and the mean is taken over classes), we refer to it as AP.
Karlinsky et~al.~\cite{karlinsky2018} released the exact episodic data and reported the AP of 56.9 without finetuning on each episode and 59.2 with finetuning.
To train our methods without looking at the test classes, we retrained the ResNet101 backbone on the remaining 786 ImageNet classes using the standard PyTorch training script.\footnote{\url{https://github.com/pytorch/examples/tree/master/imagenet}}
For the main baseline, we initialized both the detector and the retrieval system from this network and finetuned them on the images of ImageNet-LOC (the same training classes) for detection of all classes and image retrieval, respectively.
We used exactly the same code and hyperparameters as selected for the Grozi32k dataset.
The resulting method delivered the AP of \textbf{60.4} (better than the 1-shot methods of~\cite{karlinsky2018}), which confirms the strength of our baseline.\footnote{Note that it was important to retrain backbone with test classes excluded. When initialized from the standard PyTorch weights for ResNet101, which were trained on all classes, the same setting gave AP of 73.0.}
The matching based methods were not competitive at all: the sliding window baseline and OS2D V2-init gave 15.8 and 21.8 AP, respectively.
Training OS2D did not succeed and did not lead to any improvements.
We interpret such a huge difference of results as a confirmation that the settings based on the standard detection datasets (e.g., ImageNet, PASCAL VOC, COCO) are very different from Grozi2k an INSTRE showcasing the difference between instance-level vs. semantic recognition.

\section{Additional qualitative results \label{sec:extraresults}}
\begin{figure}[t]
    \begin{tabular}{@{}c@{}c@{}c@{}c@{}}
        \includegraphics[height=29mm]{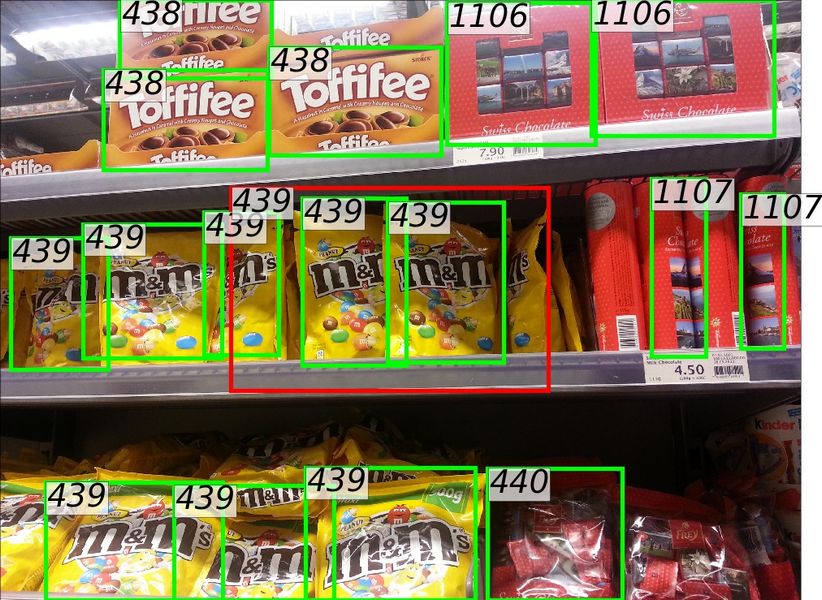} &
        \includegraphics[height=29mm]{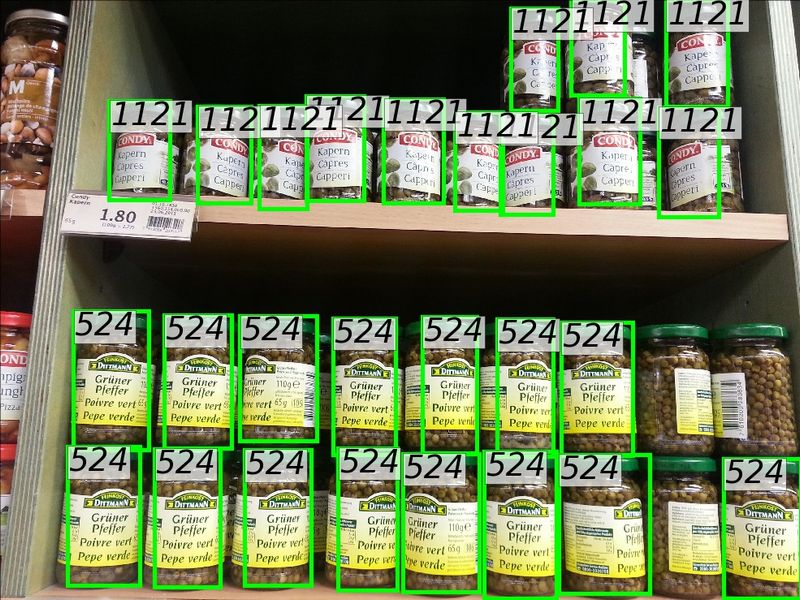} &
        \includegraphics[height=29mm]{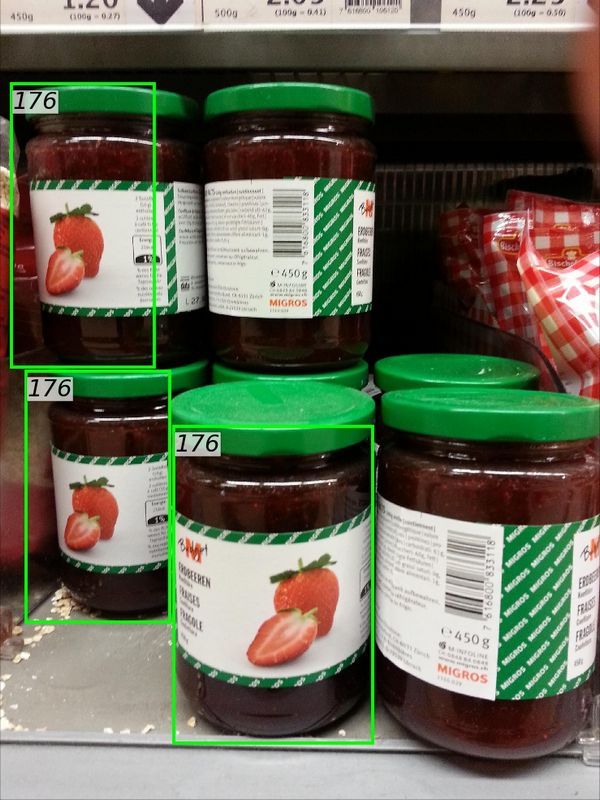} &
        \includegraphics[height=29mm]{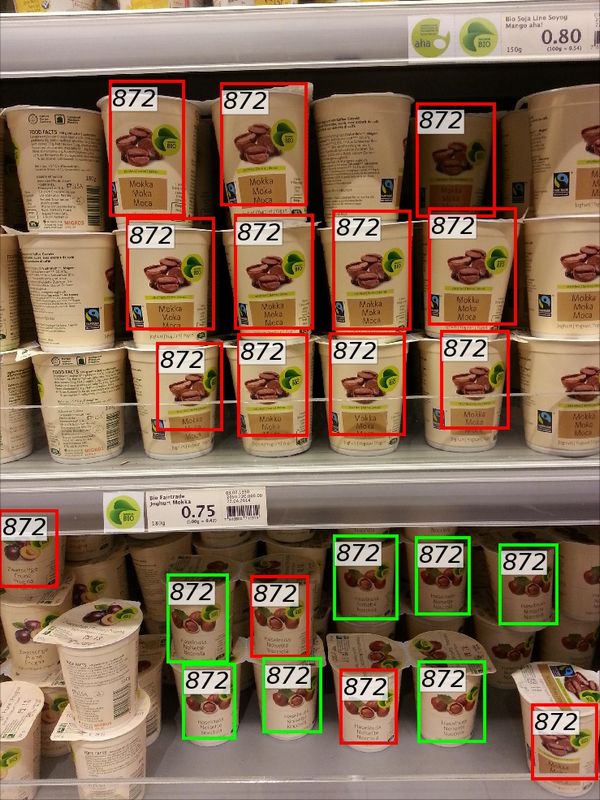}   
    \end{tabular}
    \begin{tabular}{@{}c@{}c@{}c@{}c@{}}
        \includegraphics[width=0.25\textwidth]{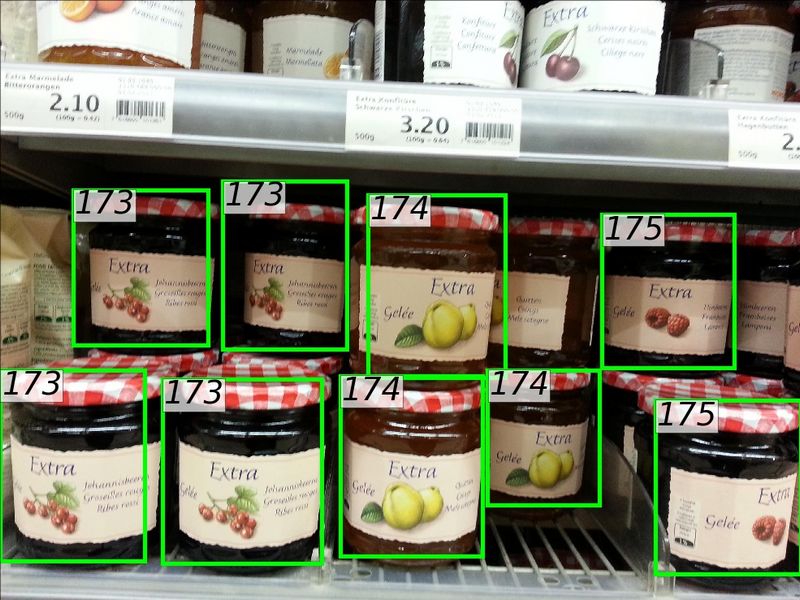} &
        \includegraphics[width=0.25\textwidth]{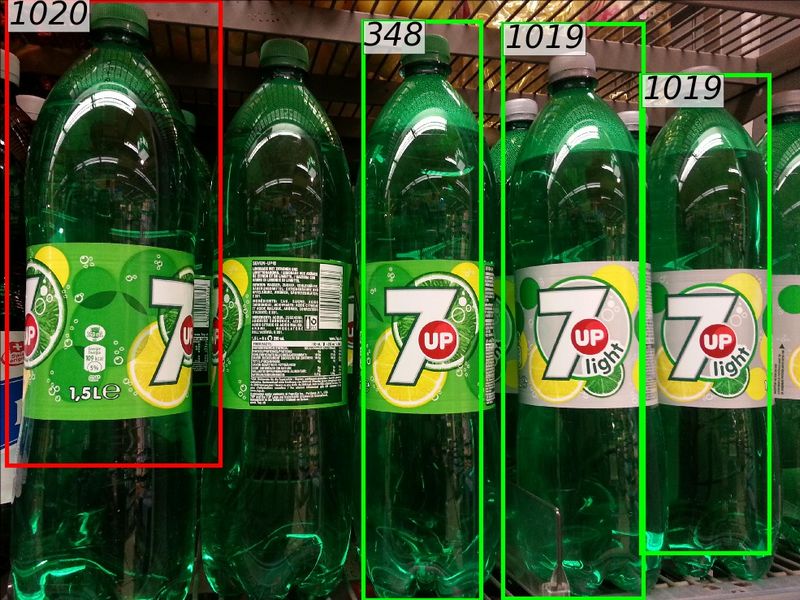} &
        \includegraphics[width=0.25\textwidth]{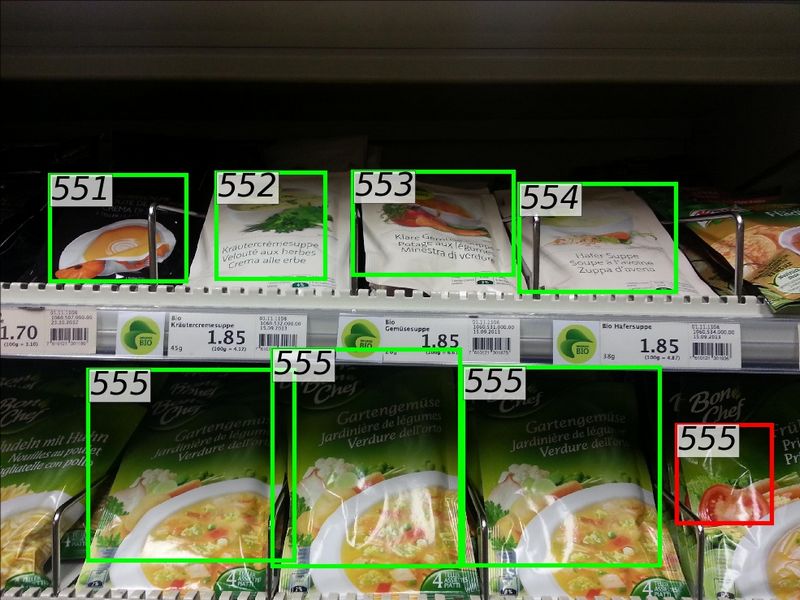} &
        \includegraphics[width=0.25\textwidth]{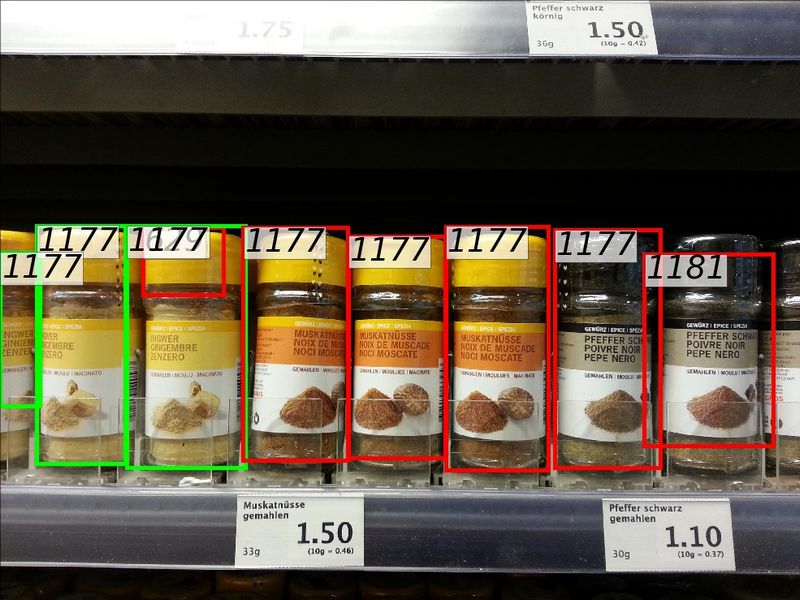}
    \end{tabular}
    \caption{Detection results on the \texttt{val-new-cl} subset of the GroZi-3.2k dataset \label{fig:res:val-new-cl}}
\end{figure}

\begin{figure}[t]
    \begin{tabular}{@{}c@{}c@{}}
        \includegraphics[width=0.5\textwidth,clip=true, trim=0mm 14mm 9mm 0mm]{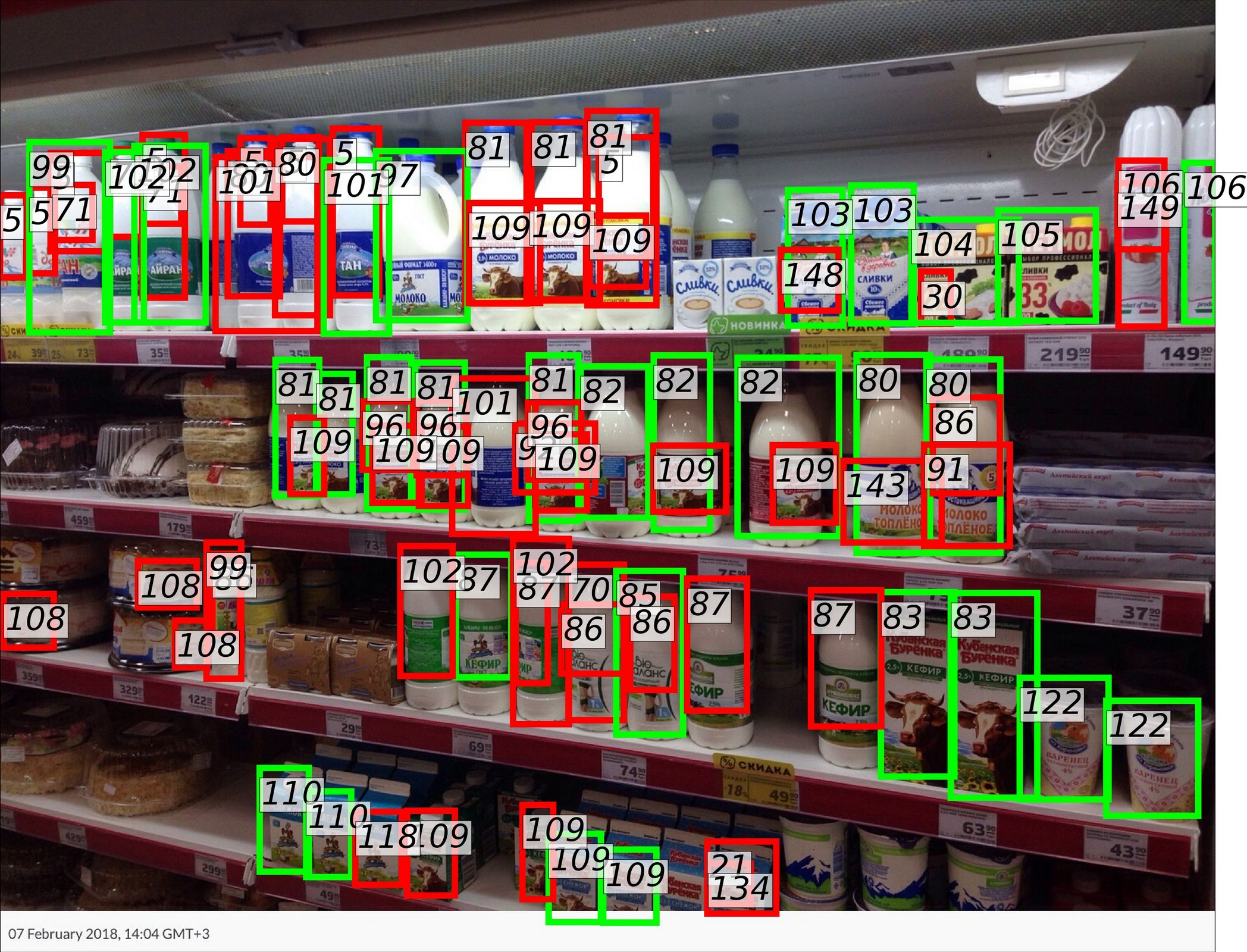} &
        \includegraphics[width=0.5\textwidth,clip=true, trim=0mm 32mm 0mm 0mm]{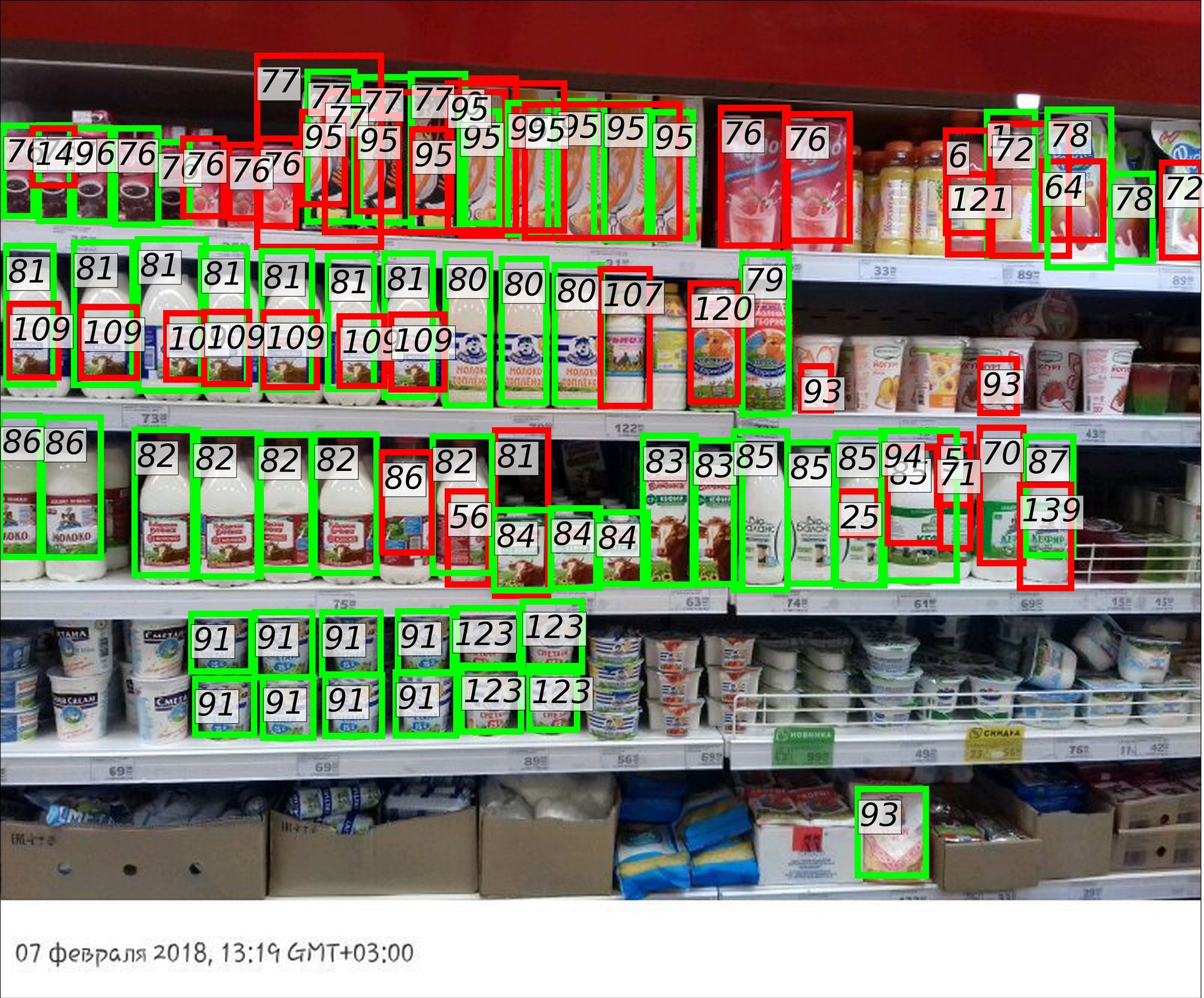} 
    \end{tabular}
    \caption{Detection results on the \texttt{dairy} test set \label{fig:res:dairy}}
\end{figure}

\begin{figure}[t]
    \begin{tabular}{@{}c@{}c@{}c@{}c@{}}
        \includegraphics[height=41mm]{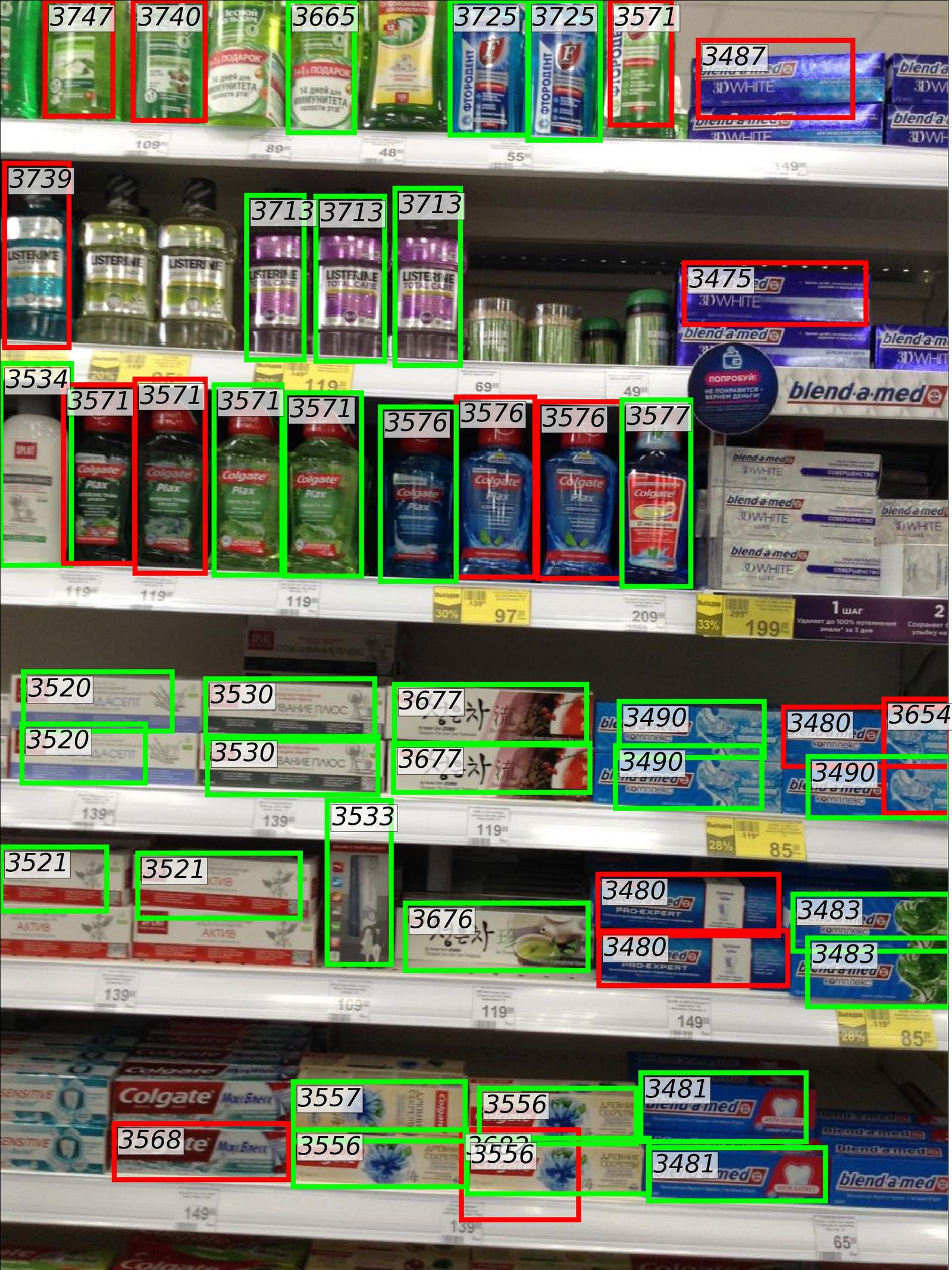} &
        \includegraphics[height=41mm]{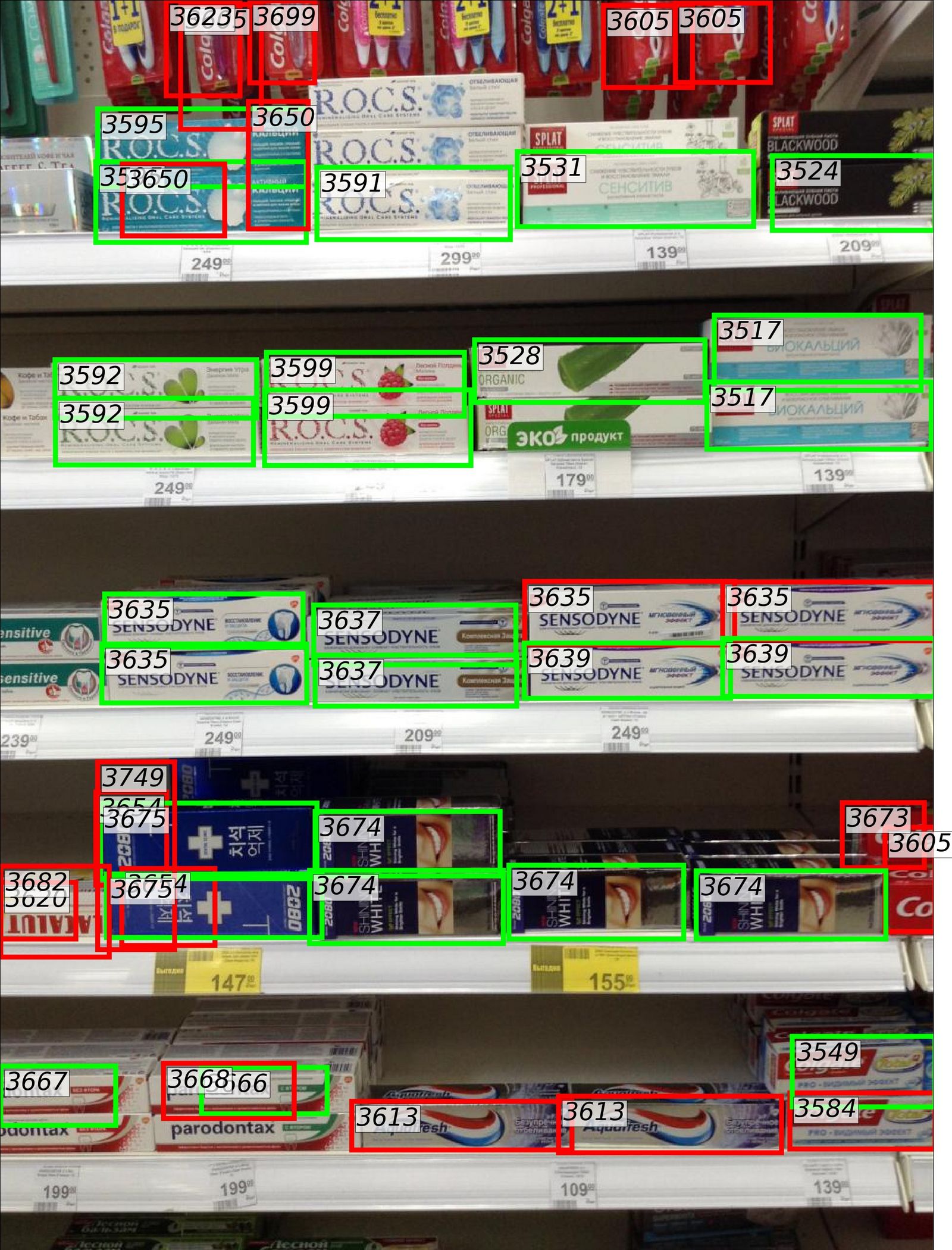} &
        \includegraphics[height=41mm]{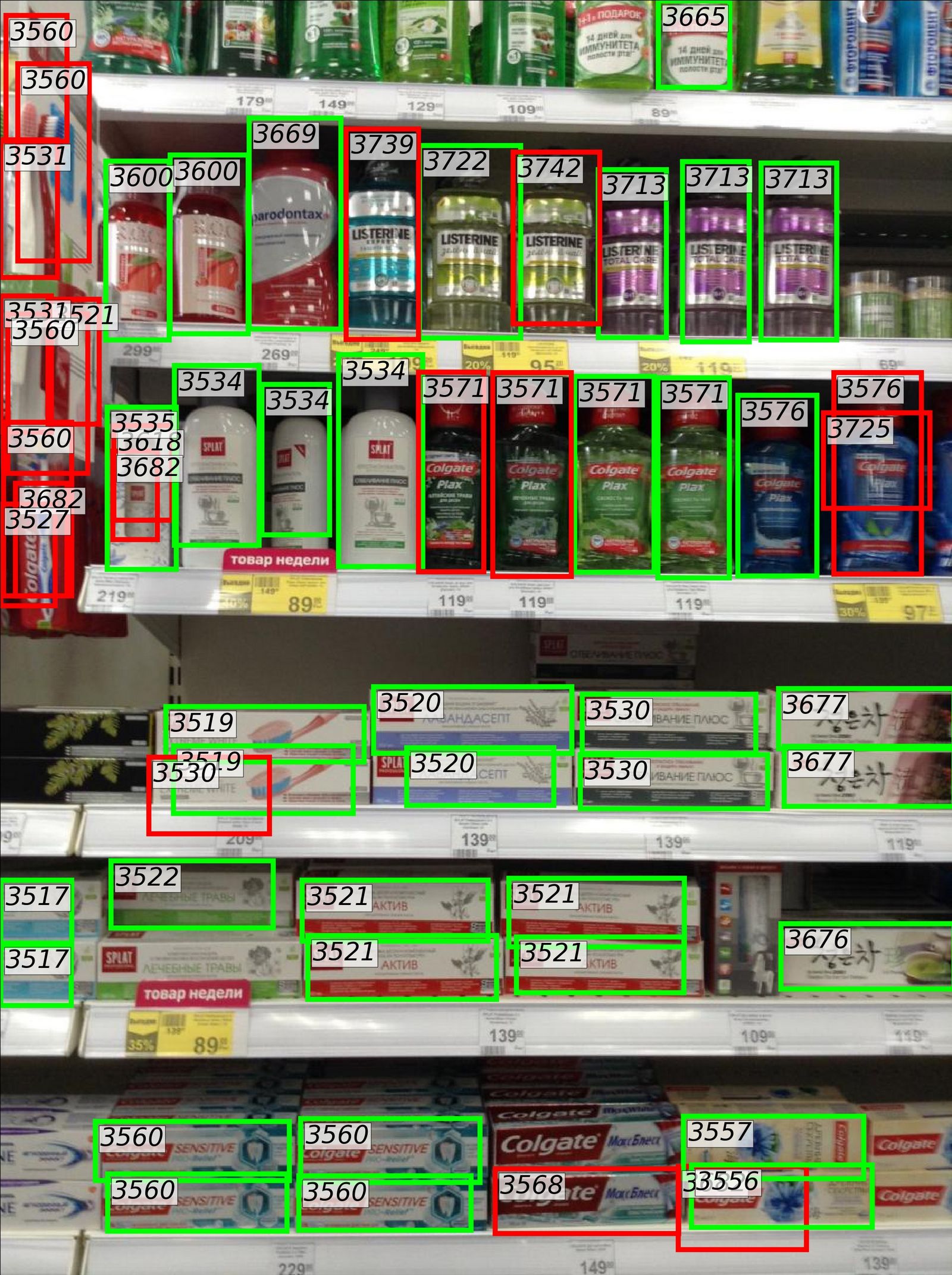} &
        \includegraphics[height=41mm]{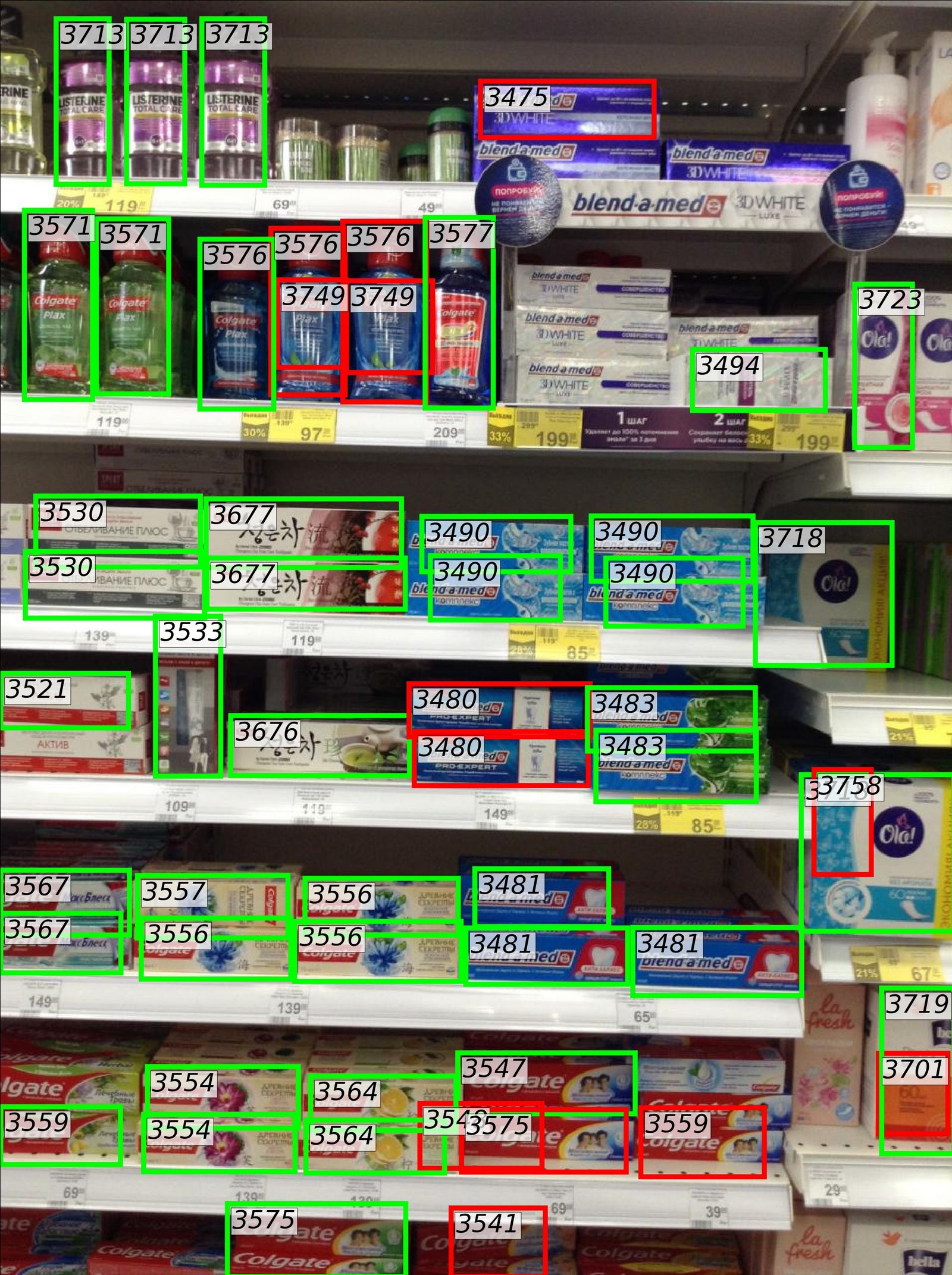}
    \end{tabular}
    \caption{Detection results on the \texttt{paste-f} test set \label{fig:res:paste-f}}
\end{figure}

In Figures~\ref{fig:res:val-new-cl}, \ref{fig:res:dairy}, \ref{fig:res:paste-f}, we present extra detection results, provided by an \methodname model.
For the purposes of visualization, we've run these results through NMS over all classes. The detection threshold was set a bit lower, so one can also see highest scoring wrong detections.

\end{document}